\documentclass[10pt,twocolumn,letterpaper]{article}

\usepackage[pagenumbers]{cvpr} 

\usepackage[utf8]{inputenc} 
\usepackage[T1]{fontenc}    
\usepackage{url}            
\usepackage{booktabs}       
\usepackage{amsfonts}       
\usepackage{nicefrac}       
\usepackage{microtype}      
\usepackage[table]{xcolor}
\usepackage{cuted}
\usepackage{caption} 
\usepackage{capt-of}
\usepackage{makecell}
\usepackage{multicol}
\usepackage{multirow}
\usepackage{graphicx}
\usepackage{pifont}

\usepackage{floatrow}     
\usepackage[tableposition=top]{caption}
\usepackage{float}
\floatstyle{plaintop}
\restylefloat{table}
\usepackage{dsfont}
\usepackage{amsmath,amssymb}
\usepackage{xspace}
\usepackage{kotex}
\usepackage[inkscapelatex=false]{svg}
\usepackage{algorithm}
\usepackage{algpseudocode}  
\usepackage{amsmath,amsfonts,bm,amssymb,mathtools,amsthm}
\usepackage{color,xcolor,xspace}
\usepackage{booktabs}
\usepackage{thm-restate}

\def\eqref#1{Eq.~(\ref{#1})}

\def\1{\bm{1}}

\def\rmI{{\mathbf{I}}}

\def\vc{{\bm{c}}}

\def\vx{{\bm{x}}}

\def\vz{{\bm{z}}}

\def\mI{{\bm{I}}}

\def\mM{{\bm{M}}}

\DeclareMathAlphabet{\mathsfit}{\encodingdefault}{\sfdefault}{m}{sl}
\SetMathAlphabet{\mathsfit}{bold}{\encodingdefault}{\sfdefault}{bx}{n}

\def\gN{{\mathcal{N}}}

\newcommand{\Ours}{VINO\xspace}%
\newcommand{\ours}{VINO\xspace}%

\definecolor{Gray}{gray}{0.9}
\definecolor{lightblue}{rgb}{0.87, 0.92, 0.97} 

\def\Plus{\texttt{+}}
\def\Minus{\texttt{-}}

\newcommand{\cmark}{\textcolor{green!60!black}{\ding{51}}} 
\newcommand{\xmark}{\textcolor{red!70!black}{\ding{55}}} 
\newcommand{\fg}{\vz_{t}^{\text{{\fontfamily{qcr}\selectfont fg}}}
\xspace}
\newcommand{\bg}{\vz_{t}^{\text{{\fontfamily{qcr}\selectfont bg}}}
\xspace}
\newcommand{\fgt}{\vz^{\text{{\fontfamily{qcr}\selectfont fg}}}
\xspace}
\newcommand{\bgt}{\vz^{\text{{\fontfamily{qcr}\selectfont bg}}}
\xspace}

\newcommand{\lbbox}[2][0.855]{%
  {\hspace*{-\fboxsep}%
    \colorbox{lightblue}{\parbox{#1\linewidth}{#2}}%
  }%
}
\newcommand{\gbbox}[2][0.855]{%
  {\hspace*{-\fboxsep}%
    \colorbox{Gray}{\parbox{#1\linewidth}{#2}}%
  }%
}
\usepackage[symbol]{footmisc}

\definecolor{cvprblue}{rgb}{0.21,0.49,0.74}
\usepackage[pagebackref,breaklinks,colorlinks,allcolors=cvprblue]{hyperref}

\def\paperID{*****} 
\def\confName{CVPR}
\def\confYear{2026}

\title{{Good Noise Makes Good Edits}: \texorpdfstring{\\}{}A Training-Free Diffusion-Based Video Editing with Image and Text Prompts}
\author{%
    Saemee Choi\thanks{Equal contribution}\\
    KAIST\\
    {\tt\small saemee99@kaist.ac.kr}
    \and
    Sohyun Jeong\footnotemark[1]\\
    KAIST\\
    {\tt\small jsh0212@kaist.ac.kr}
    \and
    Hyojin Jang\\
    KAIST\\
    {\tt\small wkdgywlsrud@kaist.ac.kr }
    \and
    Jaegul Choo\\
    KAIST\\
    {\tt\small jchoo@kaist.ac.kr}
    \and
    Jinhee Kim\\
    KAIST\\
    {\tt\small seharanul17@kaist.ac.kr}
}

\begin{document}
\maketitle

\begin{strip}
\centering
\vspace{-50pt}
\includegraphics[width=1.0\linewidth]{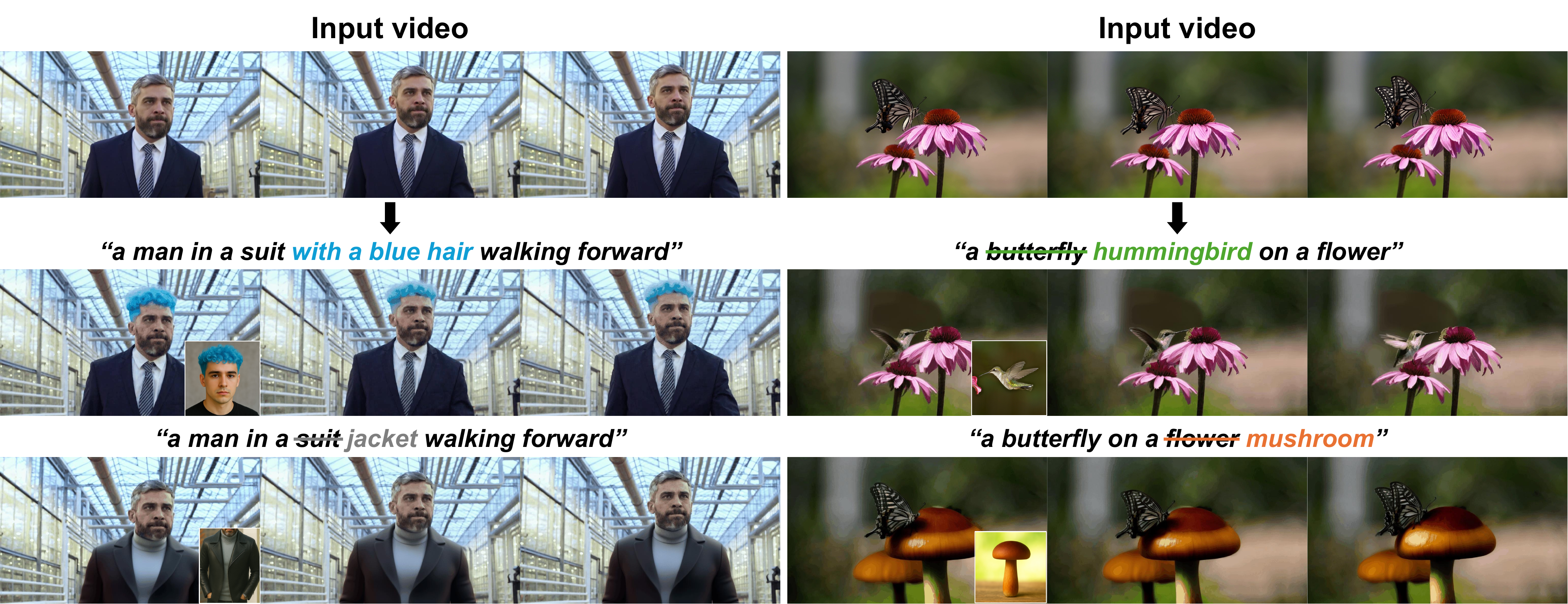}
\vspace{-20pt}
\captionof{figure}{
\textbf{Qualitative results of \ours.} Our training-free method, \textbf{\ours}, successfully achieves video edits using text and image prompts. 
}
\label{fig:teaser}
\end{strip}

\begin{abstract}
We propose \textbf{\ours}, the first \textbf{zero-shot, training-free} video editing method conditioned on both image and text. Our approach introduces $\rho$-start sampling and dilated dual masking to construct structured noise maps that enable coherent and accurate edits. To further enhance visual fidelity, we present zero image guidance, a controllable negative prompt strategy. Extensive experiments demonstrate that \ours faithfully incorporates the reference image into video edits, achieving strong performance compared to state-of-the-art baselines, all without any test-time or instance-specific training.
\end{abstract}

\section{Introduction}
\label{intro}
Diffusion-based video editing methods have shown notable achievements in recent years, enabling precise modifications of objects in videos based on user prompts~\cite{stablevideo, videdit, dreamix, revideo, controlvideo, cutandpaste}. Primarily, text-driven video editing models have gained considerable attention for their ability to modify video content while maintaining temporal consistency~\cite{tokenflow, videop2p, fatezero, tune_a_video}. These methods have demonstrated impressive results across a wide range of applications, such as appearance editing~\cite{stablevideo, videdit, magicedit, videop2p}, and style transfer~\cite{structure, tokenflow, rerenderavideo, controlvideo}.

\begin{figure*}[!t]
\centering
\includegraphics[width=1.0\linewidth]{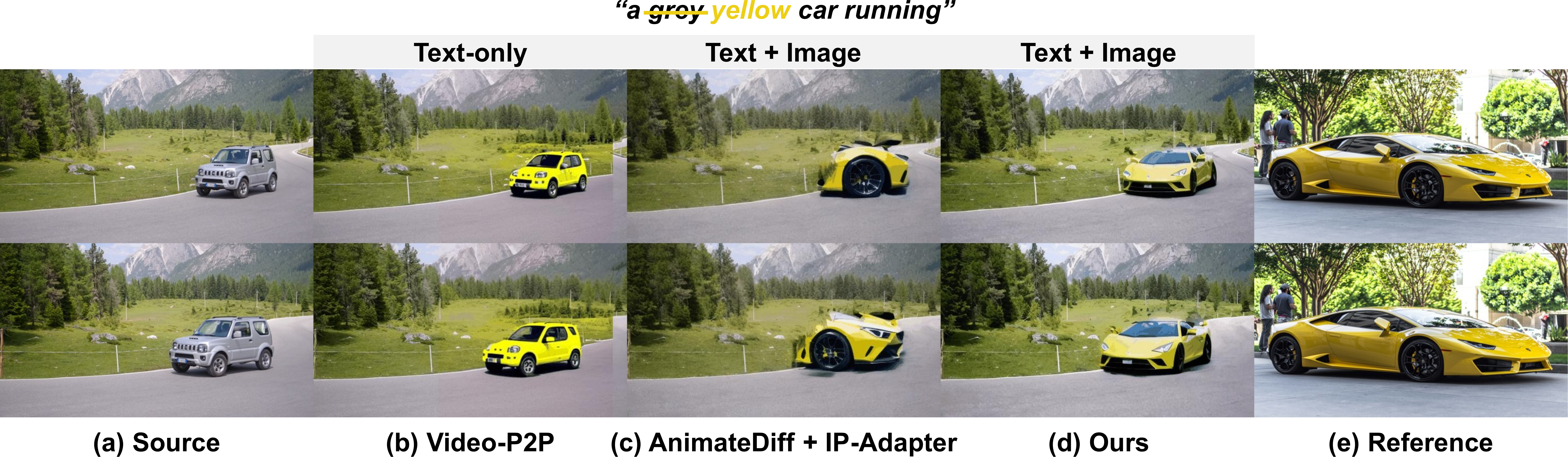}
\caption{\textbf{Importance of appropriate integration of image prompts in video editing.} Given (a) a source video, we compare (b) a state-of-the-art text-driven video editing method, (c) a naive combination of a pretrained T2V model with an image-guided module, and (d) our proposed method. The text-only approach in (b) struggles to accurately transfer the structural appearance of the car in the reference image. Simply attaching the image-guided module to a pretrained T2V model, as in (c), produces unnatural results. In contrast, our proposed training-free method in (d) precisely transfers the car in the reference image into the video with high fidelity.}
\label{fig:motiv}
\end{figure*}
\begin{figure}[!t]
\centering
\includegraphics[width=1.0\linewidth]{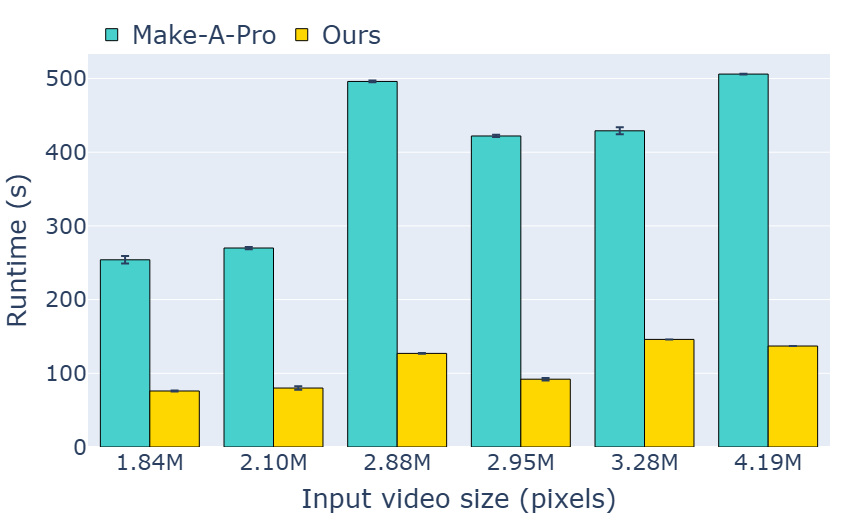}
\caption{\textbf{Runtime comparison across video resolutions.} We compare \ours with Make-A-Pro~\cite{make_a_pro}, an image-guided video editing method. Results are averaged over three runs per setting.
}\label{fig:time}
\end{figure}

However, text-based video editing has inherent limitations, particularly when modifying objects with intricate and unique details, where textual descriptions alone are often insufficient to precisely convey the desired attributes (Fig.~\ref{fig:motiv}(a)). Moreover, crafting detailed textual prompts is cumbersome, requiring extensive trial-and-error to achieve the intended modifications. This challenge becomes even more pronounced when the editing method necessitates one-shot finetuning for each edit~\cite{videop2p, fatezero, save}. 

To mitigate the need for costly text prompt engineering, recent work has explored video editing methods guided by both image and text. However, such approaches~\cite{genvideo, make_a_pro} suffer from weakened temporal consistency and require per-video finetuning, making them unsuitable for real-world editing scenarios. Other approaches employ additional guidance such as user-defined semantic keypoints alongside image and text prompts~\cite{videoanydoor, videoswap} to achieve more precise control. While effective for temporally coherent local editing guided by reference images, these methods require substantial user interaction and additional training, limiting their practicality.

Therefore, we propose \textbf{VIdeo editing with NOise control (VINO)}, the first zero-shot, training-free video editing framework conditioned on both image and text prompts. As illustrated in Fig.~\ref{fig:teaser}, our method enables localized video edits guided by reference images, without any additional training or cumbersome user interaction.

We leverage recent advances in the image domain to enable high-fidelity video editing without finetuning. In particular, pretrained image encoders and adapters~\cite{pbe, chen2024zero, stableviton, ipadapter} have shown success in injecting reference features into cross-attention layers. However, directly connecting these image-based techniques to existing text-to-video (T2V) models produces unnatural results, pasting the reference image onto the video without considering the source video’s layout (Fig.~\ref{fig:motiv}(c)). We address this semantic unnaturalness by introducing $\rho$-start sampling inspired by \cite{sdedit}, which starts DDPM sampling from intermediate time steps to generate realistic images from stroke paintings. In our strategy, both DDIM sampling and inversion begin from intermediate time steps, preserving the layout and motion of the source video.

While this strategy effectively balances the appearance of the reference and the layout of the source, it still struggles when replacing source objects with targets of significantly different sizes or shapes (Fig.~\ref{fig:ablation_mask}(a)). To overcome object size and shape discrepancies, we design \ours with a two-stage dilated dual masking strategy. In the first stage, we apply dilated masking to blend source video latents with editing latents during the sampling process, producing a coarse edited video. In the second stage, this coarse edited video is refined using dilated dual masking, enabling more flexible spatial blending for accurate appearance transfer. 
 
\begin{figure*}[!t]
\centering
\includegraphics[width=1.0\linewidth]{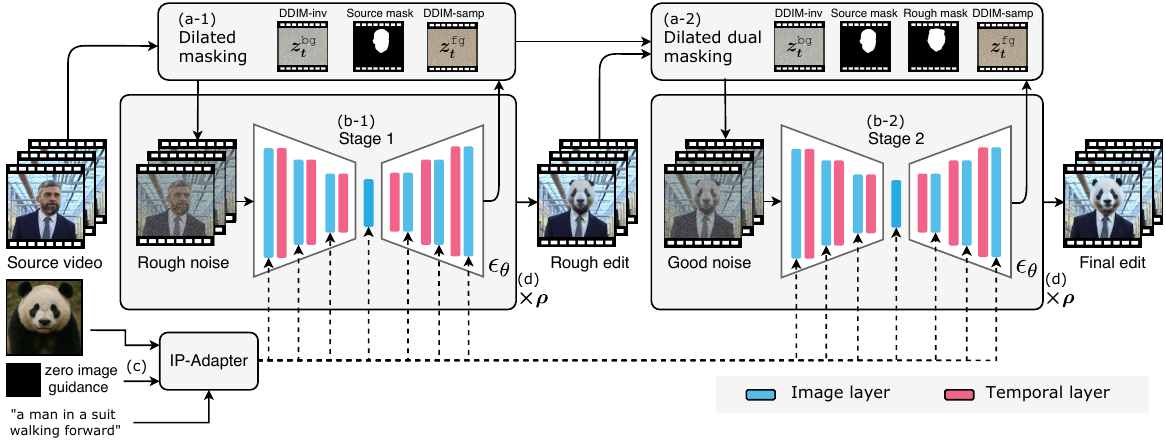}
\caption{\textbf{Overview of \ours.} Our proposed training-free approach, \ours, adopts a coarse-to-fine scheme to guide the two-stage editing process via strategically designed noise maps. Both stages leverage $\rho$-start sampling with text and image prompts. In the first stage, a rough noise map is constructed from the source video via dilated masking (a-1), which is then denoised by the pretrained T2V model to yield a rough edit (b-1). In the second stage, a refined \textit{good noise} map is derived from the rough edit via dilated dual masking (a-2). Finally, the model produces the final output based on this \textit{good noise} map (b-2), faithfully reflecting the prompts while preserving motion and structure.}
\label{fig:method2}
\end{figure*}

We further present zero-image guidance to address occasional color desaturation. This controllable negative prompt strategy steers generation away from the zero-image direction, thereby improving color fidelity. Consequently, our method achieves high visual quality and temporal consistency (Fig.~\ref{fig:motiv}(d)), while offering clear efficiency advantages over methods requiring finetuning (Fig.~\ref{fig:time}).

In summary, our key contributions are:
\begin{itemize}
\item We propose \textbf{\ours}, the first zero-shot video editing method using both text and image prompts, preserving motion and temporal consistency without training. 

\item 
We introduce a noise map construction strategy including $\rho$-start sampling to preserve the motion patterns and spatial layout of the source video, and dilated dual masking to ensure smooth appearance transfer.

\item We present zero image guidance, enhancing color saturation via controllable negative prompt blending.

\item Comprehensive quantitative and qualitative evaluations demonstrate that our method outperforms existing state-of-the-art approaches in both quality and efficiency.
\end{itemize}

\section{Methods}
We introduce \textbf{\ours}, a training-free video editing approach that performs object replacement while maintaining overall scene coherence using text and image prompts, as illustrated in Fig.~\ref{fig:method2}.
Our method is grounded in the careful integration of well-structured noise with appropriately chosen components. Specifically, we consider two key ingredients: the foreground latent, obtained via $\rho$-start sampling, and the background latent, acquired through DDIM inversion of the source video. These latents are blended using a tailored mask derived from both the source and target object masks, employing dilated dual masking to ensure smooth transitions and spatial consistency. In addition, we apply zero image guidance, a negative prompt-based strategy, to enhance color saturation and overall visual fidelity in the final output.

\subsection{Preliminaries}\label{sec:pre}
\paragraph{Latent Diffusion Models (LDMs).}
LDMs~\cite{stablediffusion} perform the diffusion process in a compressed latent space obtained from an autoencoder $\mathcal{E}, \mathcal{D}$. Given an input image \(\vx\), the encoder produces a clean latent representation \(\vz_0 = \mathcal{E}(\vx)\), where the diffusion process adds and removes noise as:
\begin{equation}
\vz_t = \sqrt{\bar{\alpha}_t}\,\vz_0 + \sqrt{1 - \bar{\alpha}_t}\,\bm{\epsilon}, \quad \bm{\epsilon} \sim \mathcal{N}(0, \rmI).
\end{equation}
To balance realism and prompt adherence, classifier-free guidance~\cite{CFG} combines conditional and unconditional diffusion model predictions as follows:
\begin{equation}
\hat{\bm{\epsilon}_\theta} = (1 + w)\bm{\epsilon}_\theta(\vz_t, t, \vc) - w\,\bm{\epsilon}_\theta(\vz_t, t, \bar{\vc}),
\end{equation}
where $w$ is the guidance scale, $\vc$ is the conditioning input, and $\bar{\vc}$ denotes the negative conditioning input. This design enables efficient training and sampling while maintaining high-quality visual fidelity consistently. 

\begin{figure*}[!t]
\centering
\includegraphics[width=0.92\linewidth]{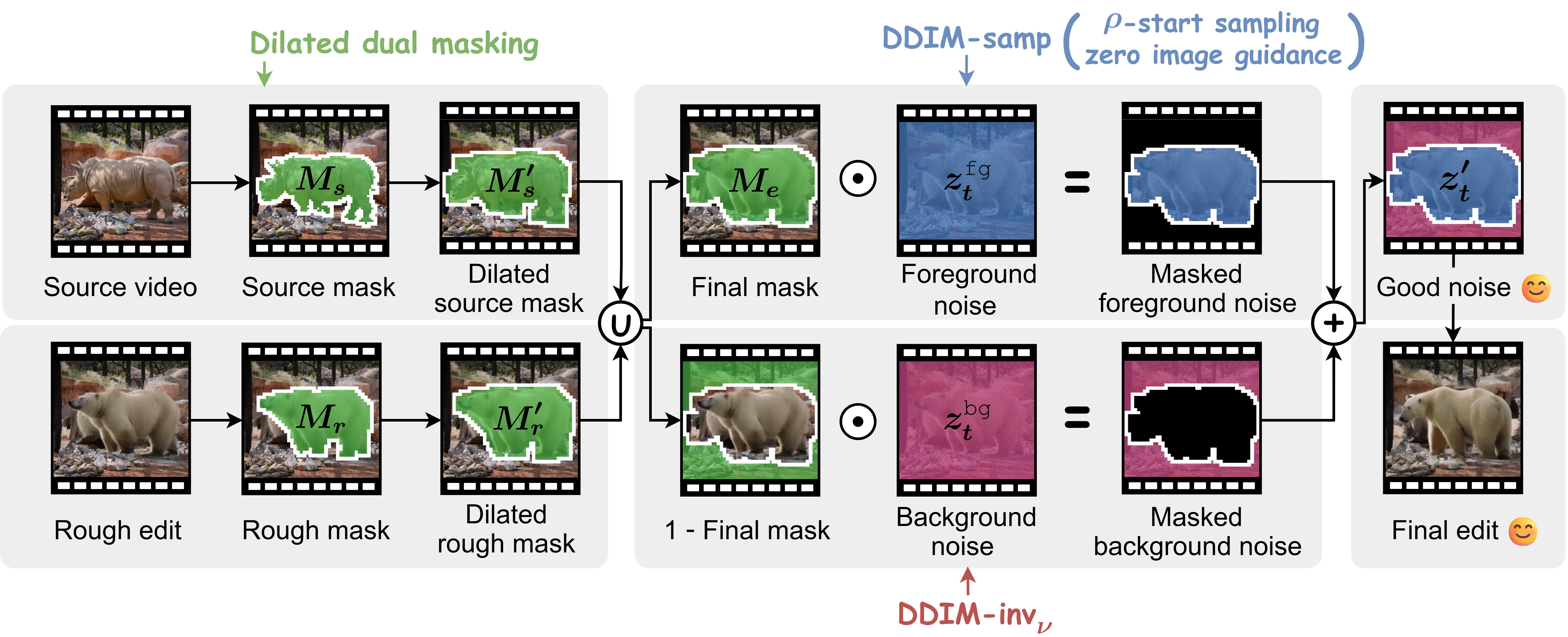}
\caption{\textbf{Controlled noise sampling for structured edits in \ours.} Foreground and background latents, obtained via $\rho$-start sampling and zero image guidance, are blended using a final mask formed by the union of dilated source and target masks. This enables clean removal of the source object and seamless insertion of the target.}
\label{fig:mask}
\end{figure*}

\paragraph{DDIM Sampling and Inversion.}
DDIM~\cite{ddim} provide a deterministic and efficient sampling process that reduces the number of steps required for faithful generation. The sampling process updates the latent as:
\begin{equation}
\begin{aligned}
    \vz_{t-1}
    &= \sqrt{\alpha_{t-1}}
    \left(
        \frac{\vz_t - \sqrt{1 - \alpha_t}\,\bm{\epsilon}_\theta(\vz_t, t)}
        {\sqrt{\alpha_t}}
    \right) \\
    &\quad + \sqrt{1 - \alpha_{t-1}} \cdot \bm{\epsilon}_\theta(\vz_t, t),
    \label{eq:ddim}
\end{aligned}
\end{equation}
where we denote this update rule as $\text{DDIM-samp}(\vz_t, t)$. The inversion process can be obtained by reversing these sampling steps, enabling reconstruction and fine-grained editing in the latent space. For editing, specific regions can be modified by blending DDIM inversion latents from the source and ongoing edit using a spatial mask:
\begin{equation}
    \vz'_{t} = \fgt_t \odot \mM + \bgt_t \odot (1 - \mM),
\end{equation}
where \(\mM\) specifies the edited region, \(\fgt_t\) and \(\bgt_t\) denote background and foreground latents, respectively. 
\subsection{Overview}
Given a source video \(\mathcal{V}_s\in\mathbb{R}^{F\times C_s \times H_s \times W_s}\), our goal is to edit the foreground object and generate an edited video $\mathcal{V}_e\in\mathbb{R}^{F\times C_s \times H_s \times W_s}$, guided by a reference image \(\mathcal{R}\in\mathbb{R}^{C_r \times H_r \times W_r} \) and a text prompt $\mathcal{T}$, while preserving the original background and motion dynamics. Here, \(F\) is the number of frames, \(C_s\) is the number of source video channels, and \(H_s\) and \(W_s\) are the height and width of each frame. Similarly, \(C_r, H_r,\) and \(W_r\) represent the channel and spatial dimensions of the reference image. 

Our approach achieves this goal through a two-stage editing pipeline, as shown in Fig.~\ref{fig:method2}. In the first stage, we construct a rough noise to enable localized object modification. In the second stage, this rough noise is refined into a more well-structured \textit{good noise}, leading to higher-fidelity editing.

\subsection{How to construct a good noise for video editing?}\label{sec:goodnoise}
We first encode the source video \(\mathcal{V}_s\) into the latent space using a pretrained VAE encoder \(\mathcal{E}\). We then initialize the noisy latent using  $\rho$-start sampling (Section~\ref{sec:noisemixing}), a modification of the DDIM sampling schedule, which enables better preservation of the source video’s structure and motion. This noisy latent is passed into a pretrained T2V diffusion model \(\epsilon_\theta\) where we incorporate both image and text guidance using the IP-Adapter~\cite{ipadapter}. The image and text features from CLIP image encoder \( \text{CLIP}_\text{img} \) and CLIP text encoders \( \text{CLIP}_\text{txt} \)~\cite{clip} are injected into each corresponding cross-attention layers of the T2V model as condition embeddings:
\begin{align}
\vc_\text{img} = \text{CLIP}_\text{img}(\mathcal{R}) \quad \text{ and } \quad  \vc_\text{txt} = \text{CLIP}_\text{txt}(\mathcal{T}),
\label{eq:condition}
\end{align}
 In addition, negative guidance using \(\bar{\vc}_\text{img}\) and \(\bar{\vc}_\text{txt}\), which denote the negative image and text embeddings respectively, helps suppress irrelevant visual or textual features.

During the denoising process with condition embeddings, we employ a foreground-background separation mechanism to perform object-centric modifications while preserving background fidelity, inspired by latent blending~\cite{latentblending,blendeddiffusion,videoswap,genvideo}. Specifically, we split the noisy latent \(\vz_t\) into foreground and background regions using the object segmentation mask \(\mM\). For the background region, we retrieve the DDIM inversion noise corresponding to the current time step to preserve the background of the source video \(\mathcal{V}_s\). In contrast, the foreground region is denoised by iteratively updating the noisy latent from the previous time step using a denoising network \(\epsilon_{\theta}\), guided by the condition embeddings. This enables the foreground object to be progressively edited to resemble the target object. Formally, the foreground latent \(\fg\) and background latent \(\bg\) at time step \(t\) are computed as:
\begin{align}
    \fgt_{t\Minus1} &= \text{DDIM-samp}(\vz'_{t}, \vc_\text{img}, \vc_\text{txt}, 
    \label{eq:ddim_samp}
    \bar{\vc}_\text{img}, \bar{\vc}_\text{txt}, t), \\
    \bgt_{t\Minus1} &= \vz_{t\Minus1}, \  \vz_{t-1} \in \{\vz_t\}_{t=1}^{\tau_\nu} \text{ from } \text{DDIM-inv}(\vz_0).
\end{align}
We then blend these two latents using \(\mM\) via our dilated dual masking (see Section~\ref{sec:doublemasking}), as follows:
\begin{align}
\vz'_{t\Minus1} = \fgt_{t\Minus1} \odot \mM + \bgt_{t\Minus1} \odot (1 - \mM),
\label{eq:mask}
\end{align}
where \( \odot \) denotes element-wise multiplication. We use different mask $M$ at each stage. Specifically, in the first stage, the dilated source mask from the source video, $\mM'_s \in\mathbb{R}^{F\times C_s \times H_s \times W_s}$, is used.
Through the denoising process in the first stage, we obtain a roughly edited video \(\mathcal{V}_r\), which reflects the general structure and appearance of the target object. However, this rough edit may still retain residual features of the source object, which should be removed for a clean visual transformation ultimately. 

To address the limitations of single mask blending, we introduce a second stage that further refines the output. In this stage, we generate the final mask \(\mM_e\) by combining the initial mask \(\mM'_s\) with the rough mask \(\mM'_r\) derived from the rough edit. Foreground latent \(\fg\) and background latent \(\bg\) are regenerated from the rough edit \(\mathcal{V}_r\) using the same procedure as in the first stage. With the final mask and the regenerated latents, we construct a more refined noise map, which we refer to as \textit{good noise}. It is designed to erase the original object and accurately follow the shape of the new object while maintaining the background. After denoising steps in the second stage, latents are decoded using the pretrained VAE decoder \(\mathcal{D}\) to generate the final output. Note that the entire pipeline is training-free, relying solely on pretrained models for mask extraction and video generation.

\subsubsection{Dilated dual masking}
\label{sec:doublemasking}
Our proposed method coordinates structured noise to define editable regions in each frame and enables selective editing of target objects (Fig.~\ref{fig:mask}). Unlike Blended Latent Diffusion (BLD)~\cite{blendeddiffusion}, which gradually shrinks masks during diffusion to address resolution degradation, our dilated dual masking uses a fixed dilation to refine boundaries and better spatially align source–target shapes.

In the first stage, only the source object mask \(\mM_s\) is available, which roughly outlines the foreground region to be edited (Fig.~\ref{fig:method2}(b-1)). However, this mask lacks contextual adaptation to the background, often resulting in sharp edges and unnatural transitions at the object boundaries.
Thus, we apply morphological dilation to smooth boundaries and facilitate seamless object replacement (Fig.~\ref{fig:method2}(a-1)). Specifically, a single-iteration morphological dilation is applied to a binary matrix \( M \) with a kernel of size \( k \times k \). For each pixel position \((p, q)\), the output value is determined as the maximum value within a neighborhood defined by \( \gN_k\), where \( \gN_k = \{ (i, j) \mid -r \leq i, j \leq r \} \) and \( r = \lfloor \frac{k}{2} \rfloor \). The dilation operation is formally expressed as:
\begin{align}
{\mM'}[p,q] = \max_{(i, j) \in \gN_k} \mM [p+i, q+j].
\label{eq:dilation}
\end{align}
As a result, the dilated source mask \(\mM'_s\) is used to blend latents in Eq.~\ref{eq:mask} to produce rough edits \(\mathcal{V}_r\).

In the second stage, we similarly obtain the dilated rough mask \(\mM'_r\) from the rough edits \(\mathcal{V}_s\) (Fig.~\ref{fig:method2}(b-2)). However, blending foreground and background latents using either the dilated source mask \(\mM'_s\) or the dilated rough mask \(\mM'_r\) poses the risk that parts of the source video may remain visible or that the generation area for the target object may not be fully covered. To address this, we leverage masks from both the first and second stages (Fig.~\ref{fig:method2}(a-2)). The final mask \(\mM_e\) is computed as the union of the dilated source mask \(\mM'_s\) and the dilated rough mask from the first stage \(\mM'_r\), as shown in the following equation below:
\begin{align}
\mM_e = \mM'_s \cup \mM'_r.
\label{eq:dual_masking}
\end{align}
Finally, we construct the \textit{good noise} $\vz'_t$ using the final mask $\mM_e$ to guide the final blending process in Eq.~\ref{eq:mask}. This enables the second stage to generate more realistic, artifact-free results, as \(\mM_e\) precisely defines the edited region. In particular, it ensures clean removal of the original object, accurate reflection of the new object's shape, and consistent preservation of the background appearance consistently.

\subsubsection{\texorpdfstring{$\mathbf{\rho}$}{rho}-start sampling}
\label{sec:noisemixing}
Preserving consistency with the source video is critical in video editing, particularly for maintaining structural integrity and coherent motion~\cite{save,camel,deco}. Inspired by SDEdit~\cite{sdedit}, we hypothesize that initiating the sampling process from a less perturbed latent, closer to the source video, helps retain both spatial structure and temporal dynamics.

To this end, we propose an alternative sampling strategy that modifies the standard DDIM schedule by starting the denoising process from a less noisy latent representation, thereby improving structural and temporal fidelity (Fig.~\ref{fig:method2}(d)).
 In the conventional reduced DDIM schedule~\cite{ddim}, sampling proceeds along a trajectory \({\vz_{\tau_1}, \ldots, \vz_{\tau_\nu}}\), starting from a highly noisy latent \(\vz_{\tau_\nu}\). In contrast, our approach begins from an earlier timestep, \(\vz_{\tau_\rho}\), where \(\rho < \nu\). This shortens the denoising trajectory to only $\rho$ steps, \( \{\vz_{\tau_1}, \ldots, \vz_{\tau_\rho}\}\), better preserving source content overall.

The parameter \(\rho\) controls how early the sampling begins in the reverse diffusion process. A smaller \(\rho\) initiates from a latent closer to the source video, enabling better preservation of motion patterns and spatial layout, such as object direction and position. Conversely, a larger \(\rho\) facilitates greater transformation, allowing the edited video to more closely resemble the reference image in terms of object appearance.
By adjusting $\rho$, we gain fine-grained control over the degree of signal retention, enabling higher fidelity in both motion and overall scene consistency.

\begin{table*}[!t]
\centering
\caption{\textbf{Quantitative comparison.} We compare \ours against four state-of-the-art models and two naive adaptations of image editing and video generation models for video editing. We measure performance using both automatic metrics and human evaluations to assess text, image, and motion alignment, temporal consistency, and overall quality.}

\label{table:combined}
\resizebox{1.0\textwidth}{!}{
\begin{tabular}{l>{\centering\arraybackslash}p{1.3cm}ccccccccccc}
\toprule
{\multirow{2.4}{*}{{\textbf{Method}}}} & \multicolumn{3}{c}{\textbf{Model support}} &
 \multicolumn{4}{c}{{\textbf{Model scoring metrics}}} &
\multicolumn{5}{c}{{\textbf{Human evaluations}}} \\
 \cmidrule(l{2pt}r{2pt}){2-4} \cmidrule(l{2pt}r{2pt}){5-8} \cmidrule(l{2pt}r{2pt}){9-13}
& Text & Image& Train-free & DINO $\uparrow$   & CLIP-T $\uparrow$  & Temp $\uparrow$ & PSNR $\uparrow$ & Image $\downarrow$  & Text $\downarrow$ & Motion $\downarrow$ &Temp $\downarrow$ & Overall $\downarrow$ \\
\midrule
 SDEdit-Video\textsuperscript{\textdagger}~\cite{sdedit} & \cmark & \xmark & \cmark &0.431&0.303&0.959& 16.3&4.89&5.08&5.93&5.02&5.36\\
  FateZero~\cite{fatezero} & \cmark & \xmark & \cmark &0.438&0.311&0.967& 20.1&4.59&4.53&4.44&5.06&4.75\\
 Video-P2P~\cite{videop2p}  & \cmark & \xmark & \xmark &0.424&0.305&0.971& 19.0&4.00&4.07&3.68&4.09&3.70\\
 TokenFlow~\cite{tokenflow} & \cmark & \xmark & \cmark &\underline{0.480}&\textbf{0.325}&\textbf{0.975}& 19.6&\underline{3.78}&\underline{3.09}&\underline{2.61}&\underline{2.50}&\underline{2.95}\\
\midrule
 AnimateDiff~\cite{animatediff}\Plus IPA~\cite{ipadapter} & \cmark & \cmark & \cmark &0.399&0.308&0.955& \textbf{24.7}&4.59&5.01&5.24&4.30&4.93\\
 Make-A-Pro~\cite{make_a_pro}  & \cmark & \cmark & \xmark &0.444&0.309&0.965& 16.9&3.93&4.24&3.64&5.00&4.35\\
 \rowcolor{Gray}\textbf{\Ours \ ({Ours})} &\cmark  & \cmark & \cmark  &\textbf{0.498}&\underline{0.321}&\underline{0.972}& \underline{23.4}&\textbf{2.22}&\textbf{1.98}&\textbf{2.46}&\textbf{2.03}&\textbf{1.96}\\
\bottomrule
\end{tabular}
}
\end{table*}

\begin{figure*}[!t]
\centering
\includegraphics[width=1.0\linewidth]{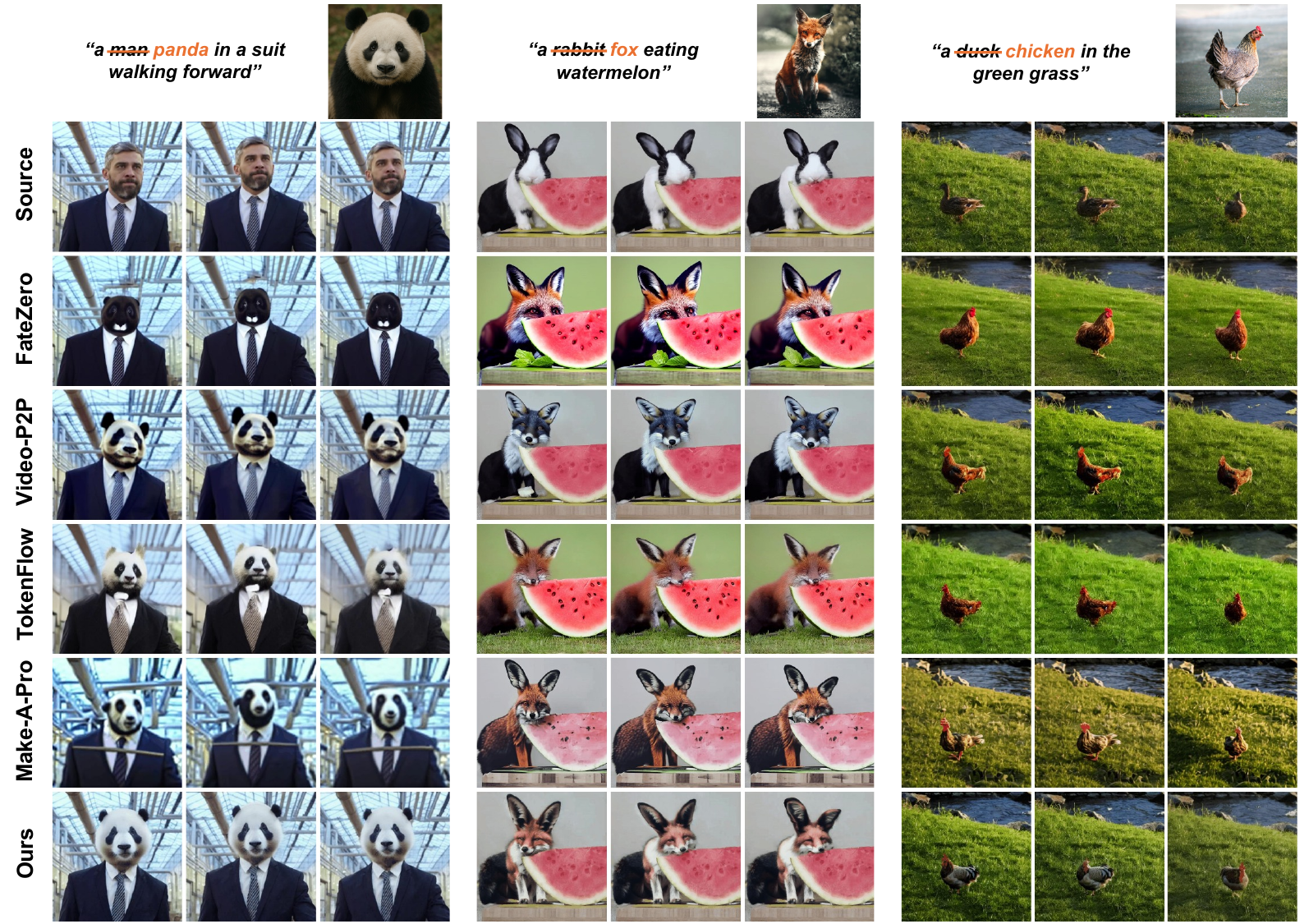}
\caption{\textbf{Qualitative comparison.} We compare the generation quality of \ours against four state-of-the-art baseline models. \ours better preserves reference appearance across frames, while baselines struggle to retain key visual features of the target object.
}
\label{fig:mulref}
\end{figure*}

\subsection{Zero image guidance}
\label{sec:ane}
To enhance color fidelity, we present zero image guidance (Fig.~\ref{fig:method2}(c)). It blends the conventional negative image embeddings~\cite{CFG, glide}, defined by a zero vector of the same size as the CLIP image embedding, with a zero image guidance, which refers to the CLIP image embedding derived from an all-zero image. The final negative image guidance \(\bar{\vc}_\text{img}\) is computed as a weighted sum of the two, controlled by a blending factor $\gamma$:
\begin{equation}
\bar{\vc}_\text{img} = \gamma \cdot \text{CLIP}_\text{img}(\rmI_0) + (1-\gamma) \cdot \mathbf{0},
\label{eq:ane_gamma}
\end{equation}
where \( \gamma \in [0, 1] \) is a hyperparameter,  
\( \mathbf{0} \in \mathbb{R}^d \) is a zero vector with the same size as the CLIP image embedding,  
\( \text{CLIP}_\text{img}(\cdot) \) is the CLIP image encoder,  
and \( \rmI_0 \in \mathbb{R}^{C \times H \times W} \) is an all-zero image. This formulation offers a parameterized trade-off between the two types of negative guidance, allowing finer control over the color intensity of the generated object. The resulting \(\bar{\vc}_\text{img}\), combined with the conventional \(\bar{\vc}_\text{txt}\), is used as negative guidance during the sampling process in Eq.~\ref{eq:ddim_samp}.

\begin{figure*}[!t]
\centering
\includegraphics[width=1.0\linewidth]{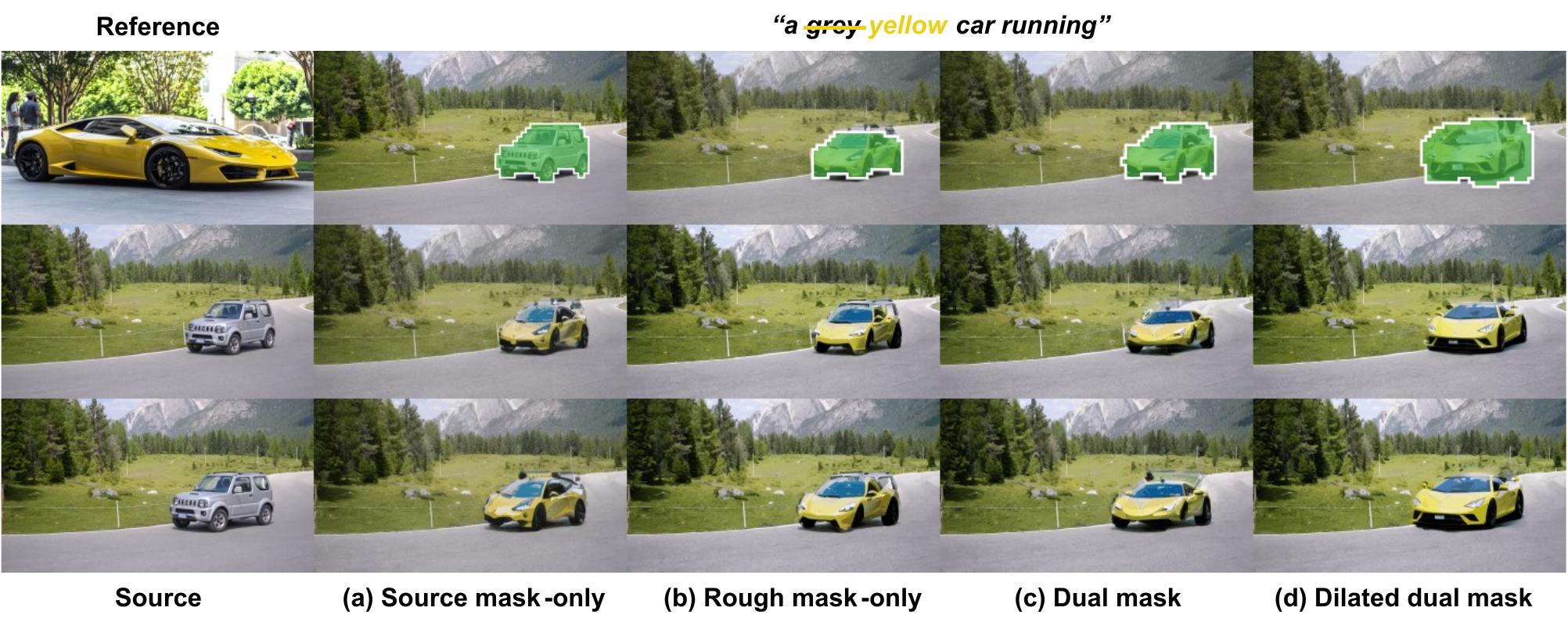}
\caption{\textbf{Ablation study on the dilated dual masking.} The output of the first stage using the source video \(\mathcal{V}_s\) and source object mask is shown in (a). Subsequently, the outputs of the second stage from the intermediate video \(\mathcal{V}_r\) using different masking strategies are presented in (b)-(d). Masks are overlaid on the corresponding video. The proposed dilated dual masking (d) achieves the most accurate edits.}
\label{fig:ablation_mask}
\end{figure*}

\begin{figure}[!t]
    \centering
\includegraphics[width=1\linewidth]{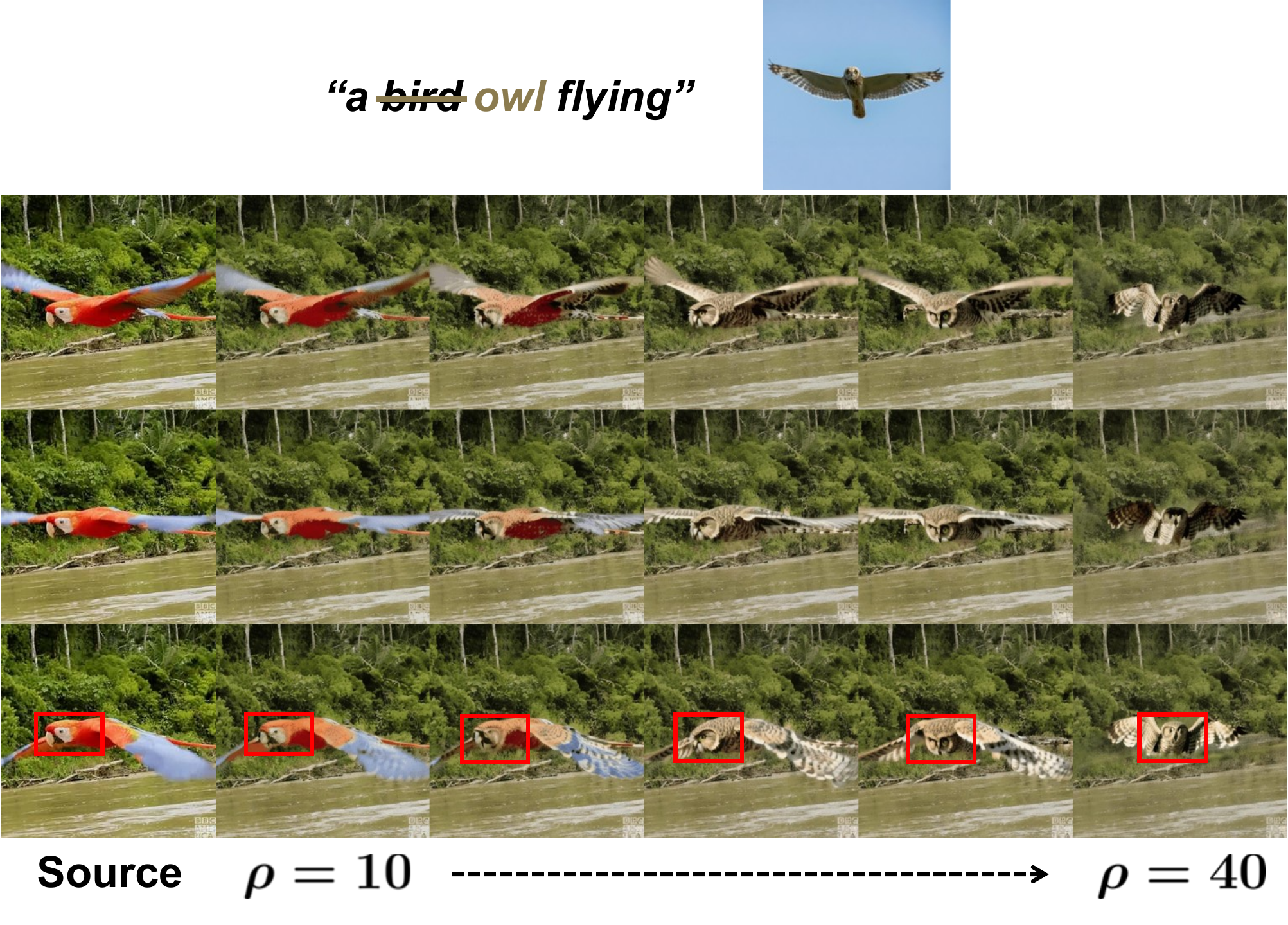}
\caption{\textbf{Ablation study on the effect of varying the \(\rho\)\ value.} }
\vspace{-2mm}
\label{fig:clean_latent}
\end{figure}

\section{Experiments}

\noindent
\textbf{Dataset.} 
We evaluate \ours on 20 video-image pairs. Videos are sourced from DAVIS~\cite{davis}, Video-P2P~\cite{videop2p}, and a curated set covering various object categories (e.g., humans, animals, everyday items) to ensure broad semantic and visual diversity. Reference images include both real and synthetic samples generated by models such as GPT-4o and Gemini.

\noindent
\textbf{Baseline models.} 
We compare \ours against six representative video editing methods, as summarized in Table~\ref{table:combined}, which use text or image prompts for semantic appearance control.
{SDEdit-Video}\footnote[2]{We use the HuggingFace implementation of SDEdit extended to video.}~\cite{sdedit}, {FateZero}~\cite{fatezero}, {Video-P2P}~\cite{videop2p}, and {TokenFlow}~\cite{tokenflow} represent text-guided video editing methods.
To our knowledge, {Make-A-Protagonist (Make-A-Pro)}~\cite{make_a_pro} is the only publicly available method that supports both image and text conditioning, making it the most directly comparable to \ours.
{AnimateDiff~\cite{animatediff}} \Plus {IP-Adapter (IPA)~\cite{ipadapter}} serves as a modular baseline that integrates a pretrained T2V model with an image-guided adapter, illustrating the limitations of naive fusion strategies.

\noindent
\textbf{Implementation details.}
\label{sec:implementation}
Our main experiments adopt AnimateDiff~\cite{animatediff} as the base T2V model, IP-Adapter v1.5~\cite{ipadapter} for reference image conditioning, and SAM2~\cite{sam2} for mask generation. All experiments are conducted on a single NVIDIA A100 GPU. Additional implementation details, including hyperparameters, baseline setups, and extensions to other T2V backbones, are provided in Appendix~\ref{appen:implement}.

\noindent
Due to space constraints, we present representative qualitative comparisons and ablation results in the main paper, with full qualitative results provided in Appendix~\ref{appen:frames}.

\begin{figure}[!t]
\centering
\includegraphics[width=1.0\linewidth]{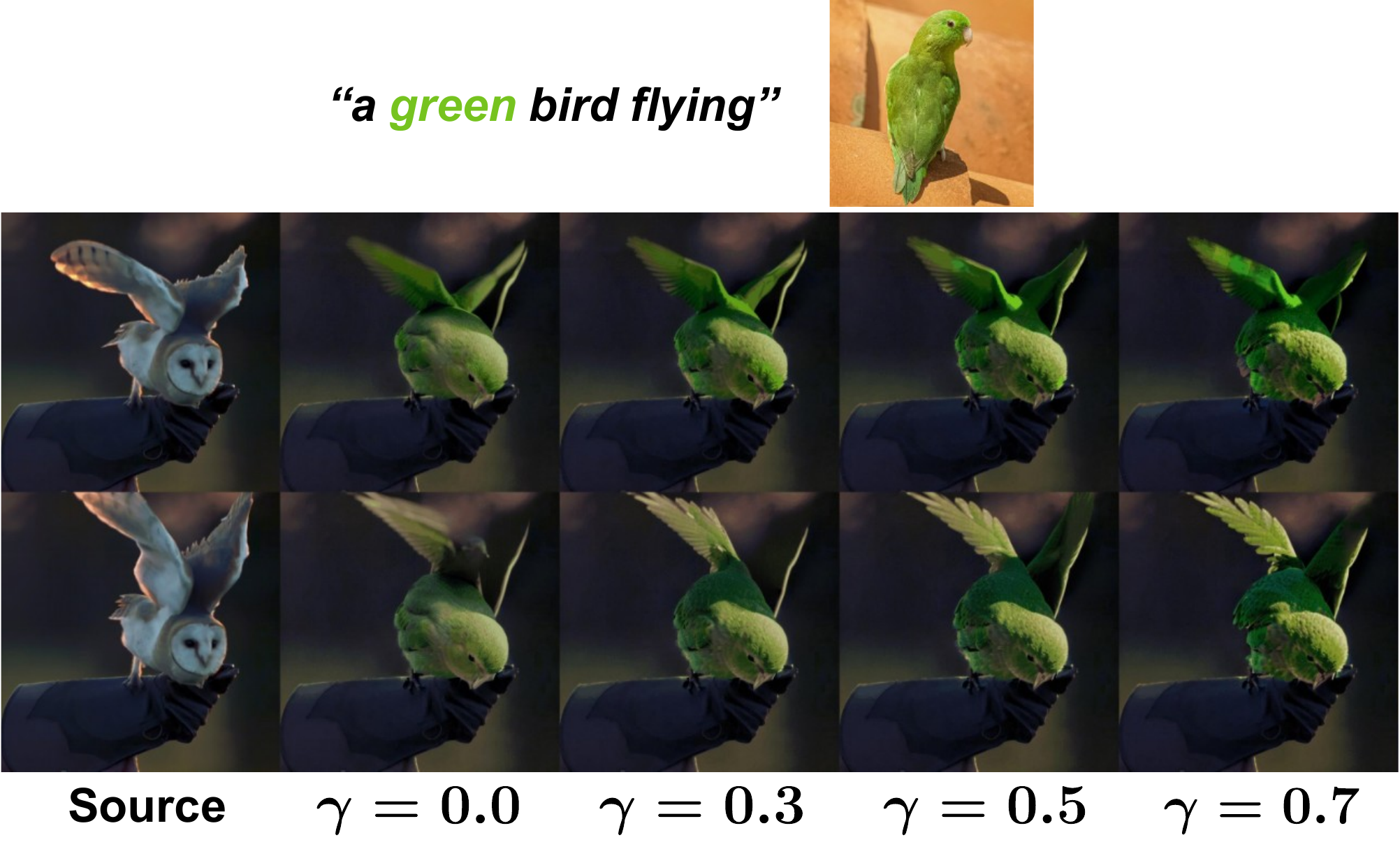}
\caption{\textbf{Ablation study on the image guidance scale $\gamma$.}}
\vspace{-2mm}
\label{fig:ablation_gamma}
\end{figure}

\subsection{Comparisons with state-of-the-art methods}
\label{sec:experiment}

\noindent
\textbf{Quantitative results.} 
We evaluate \ours using four metrics: DINO score measures visual alignment via cosine similarity between features of the reference image and generated video~\cite{dino}. 
CLIP-Text score quantifies text alignment by computing cosine similarity between CLIP embeddings of the generated video and the text prompt~\cite{clip}. Temporal score assesses motion consistency through cosine similarity between CLIP image embeddings of adjacent frames~\cite{clip}. PSNR score measures how accurately the background of the source video is preserved by computing pixel-level similarity between the generated and source frames.

\ours achieves the highest DINO score among all baselines, indicating strong visual fidelity to the reference appearance. In CLIP-Text and temporal scores, it performs comparably to TokenFlow while outperforming other methods. Notably, \ours is compatible with existing methods such as TokenFlow and can serve as an additional appearance control module. This composition yields further performance gains, as detailed in the Appendix~\ref{appen:applicability}. For PSNR, AnimateDiff+IPA and \ours noticeably outperform the other methods. The small gap between them, despite both using the same latent blending strategy, arises from the edited area. AnimateDiff+IPA modifies a comparatively smaller region, whereas \ours employs dilated dual masking, which introduces a larger generation area, leading to a marginally lower PSNR.

\noindent
\textbf{User study.} We conduct a user study with 25 participants to assess perceptual quality across five criteria: image alignment, text alignment, motion alignment, temporal consistency, and overall quality. \Ours consistently outperforms all baselines across all criteria, demonstrating strong perceptual fidelity in both appearance and motion.

\noindent
\textbf{Qualitative results.} 
Qualitative comparisons, shown in Fig.~\ref{fig:mulref}, further demonstrate the effectiveness of \ours in preserving object appearance and motion across various editing scenarios, consistently outperforming baselines.

In the man-to-panda translation, baselines either fail to generate a panda or produce artifacts or distortions, while \ours yields a clean and consistent result. A similar trend appears in the duck-to-chicken translation, where most methods generate generic chicken-like color rather than matching the reference, whereas \ours preserves its specific visual traits. In the rabbit-to-fox translation, most baselines struggle to maintain appearance fidelity or temporal coherence. Make-A-Pro, in particular, suffers from temporal flickering, including disappearing or reappearing limbs. In contrast, \ours maintains the correct fox appearance while preserving the background, producing coherent structure and stable motion throughout the video.

\begin{table}[t]
\caption{\textbf{Comprehensive ablation study of \ours.} We evaluate the contribution of each component to the overall performance of our method: (a) base model, (b) $\rho$-start sampling, (c) dual masking, (d) dilation, and (e) zero image guidance.}
\label{table:abl_quan}
\begin{center}
\resizebox{0.9\textwidth}{!}{%
\begin{tabular}{lccc}
\toprule
 {Method} & 
DINO $\uparrow$   & CLIP-T $\uparrow$  & Temp $\uparrow$ \\
\midrule
(a) & 0.399&0.308&\underline{0.955}\\
(a) + (b) &0.429&0.311&0.953\\
(a) + (b) + (c) &0.466&\underline{0.316}&0.954\\
(a) + (b) + (c) + (d) &\underline{0.474}&\underline{0.316}&\underline{0.955}\\
\rowcolor{Gray} (a) + (b) + (c) + (d) + (e) &\textbf{0.498}&\textbf{0.321}&\textbf{0.972}\\
\bottomrule
\end{tabular}
}
\end{center}
\end{table}

\subsection{Ablation study}
\noindent
\textbf{Effectiveness of dilated dual masking.}
We conduct an ablation study on our dilated dual masking by progressively applying different masking schemes (Fig.~\ref{fig:ablation_mask}). 
Using only the source mask constrains the generation to the contours of the source object, often resulting in unrealistic shapes and artifacts when the source and target differ significantly (Fig.~\ref{fig:ablation_mask}(a)). 
Applying only the rough mask alleviates this issue but leaves visible remnants of the source object (Fig.~\ref{fig:ablation_mask}(b)). Taking the union of both masks suppresses these remnants but introduces boundary artifacts and limits the detail due to the rigid mask constraint (Fig.~\ref{fig:ablation_mask}(c)).

In contrast, our dilated dual masking effectively mitigates these limitations by dilating both masks before fusion, enabling the generated object to extend beyond the strict union region. This flexibility promotes more natural shape adaptation and smoother blending with the surrounding context (Fig.~\ref{fig:ablation_mask}(d)). As a result, our dilated dual masking achieves both structural coherence and visual realism.

\noindent
\textbf{Effectiveness of $\rho$-start sampling.}
As shown in Fig.~\ref{fig:clean_latent}, adjusting \(\rho\), which controls the sampling start point, enables a trade-off between target appearance and source motion. Larger \(\rho\) values cause the model to deviate from the source motion and layout. For example, the parrot facing left in the source video becomes an owl facing forward in the edited result. Smaller \(\rho\) values better preserve the source motion and spatial dynamics, producing an owl that correctly faces sideways. However, overly small \(\rho\) weakens the transformation, leading to incomplete replacement of the source object. An appropriate choice of \(\rho\) achieves a balanced edit, maintaining both appearance fidelity and motion preservation.

\noindent
\textbf{Effectiveness of zero image guidance.}
We analyze the impact of \(\gamma\) in our zero image guidance. As shown in Fig.~\ref{fig:ablation_gamma}, increasing $\gamma$, which scales the strength of the zero image guidance (see Eq.~(\ref{eq:ane_gamma})), leads to more saturated colors. This suggests that incorporating the zero image guidance into the negative image effectively enhances color intensity, providing a simple yet powerful control for visual appearance.

\noindent
\textbf{Quantitative results.} Table~\ref{table:abl_quan} summarizes the effect of each component in our method. 
As we progressively add components (a)–(e), DINO and CLIP-Text scores increase monotonically, indicating that the proposed modules jointly improve appearance transfer from the reference image as well as adherence to the text prompt. In contrast, the temporal score remains relatively stable for variants (a)–(d), but improves significantly once zero image guidance (e) is introduced, leading our complete model to achieve the best performance across all metrics. These results quantitatively demonstrate that each component contributes effectively to temporally coherent video editing conditioned on both image and text.

\section{Conclusion}
In this work, we propose \ours, a novel zero-shot, training-free video editing method guided by both image and text prompts. It enables accurate object replacement with strong temporal coherence, outperforming state-of-the-art methods in visual quality and motion consistency, and efficiency.

{
    \small
    \bibliographystyle{ieeenat_fullname}
    \bibliography{main}
}

\newpage
\clearpage

\section*{Appendix}
This supplementary material complements the main manuscript, providing additional experimental results and implementation details that could not be included because of space limitations.

\begin{itemize}
\item \textbf{Appendix~\ref{appen:applicability}} illustrates the applicability of our method across various text-to-video models.
\item \textbf{Appendix~\ref{appen:frames}} includes complete qualitative comparison and ablation studies.
\item \textbf{Appendix~\ref{appen:relatedwork}} provides related work.
\item \textbf{Appendix~\ref{appen:limit}} discusses the limitations of our method.
\item \textbf{Appendix~\ref{appen:exp_settings}} details additional experimental settings.
\item \textbf{Appendix~\ref{appen:alg}} presents the overall algorithm of our method.
\item \textbf{Appendix~\ref{appen:qual}} showcases additional qualitative results.
\end{itemize}

\appendix

\section{Applicability to various T2V models}
\label{appen:applicability}

As shown in Table~\ref{table:supple_plug}, our method is compatible with various text-to-video (T2V) models, enabling training-free video editing conditioned on both images and text. We evaluate the compatibility of our approach by extending the text-based generative model AnimateDiff~\cite{animatediff} and the text-based editing model TokenFlow~\cite{tokenflow} with our proposed components, including dilated dual masking, $\rho$-start sampling, zero-image guidance, and the image-guided module. 

Integrating our method into AnimateDiff yields higher scores across most metrics, demonstrating substantial improvements in structural consistency and visual similarity to the reference image. When applied to TokenFlow, DINO and PSNR also improve, indicating the integration strengthens reference alignment even in the text-based editing models. 

However, our method shows a lower temporal consistency score than that of text-prompt-only baselines. This reflects a trade-off inherent to image-based conditioning, where aligning each frame to a high-fidelity reference image introduces larger pixel-level variations, even though the edits remain perceptually coherent. Despite this trade-off, the consistent gains achieved with only marginal drops in temporal consistency across both generation and editing models underscore the applicability and generality of our method.

\begin{figure*}[!t]
\centering  
\includegraphics[width=1.0\linewidth]{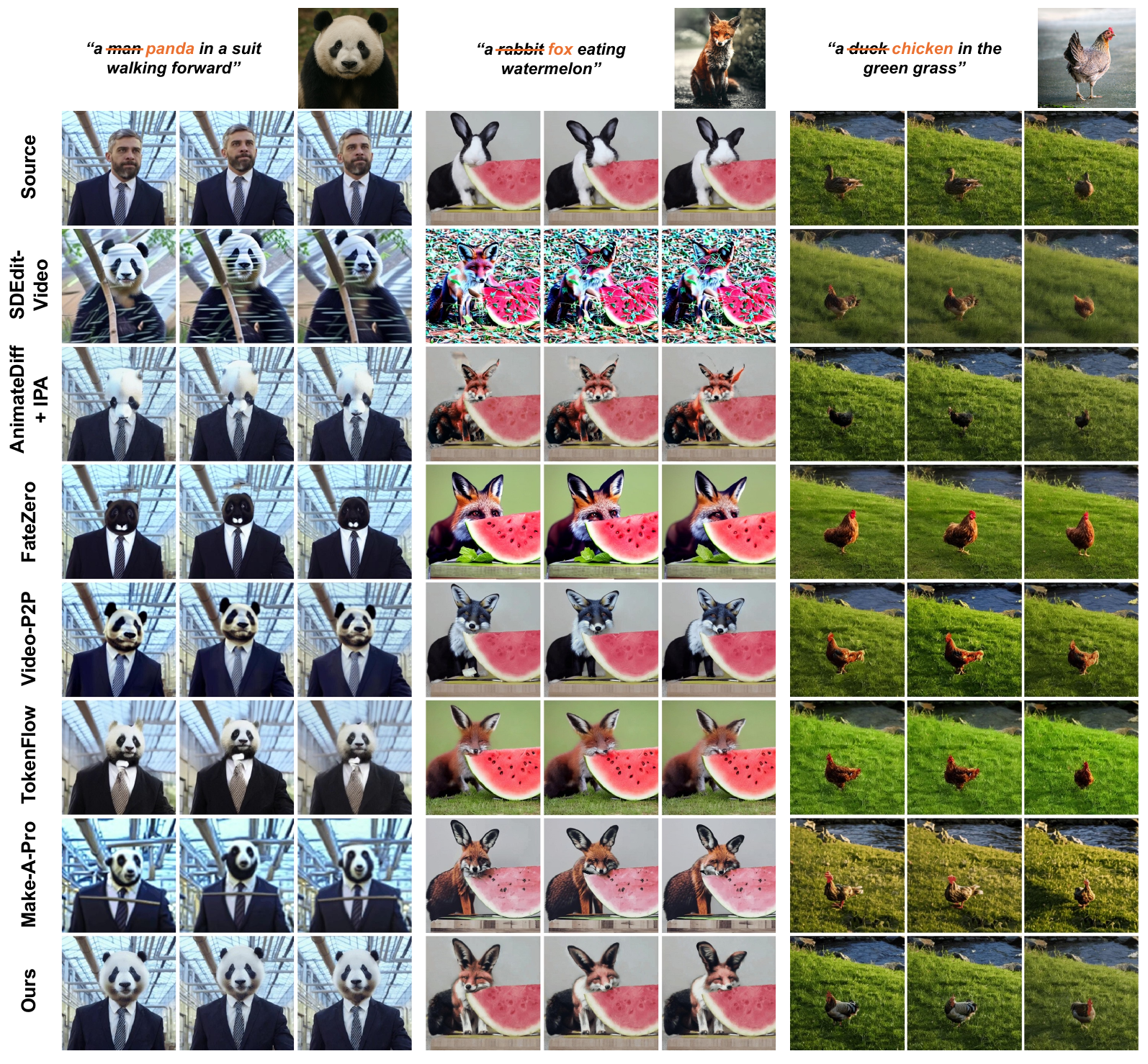}
\caption{\textbf{Full qualitative comparison with baselines.} We provide an extended version of Fig.~\ref{fig:mulref} from the main manuscript by including results from SDEdit-Video~\cite{sdedit} and AnimateDiff~\cite{animatediff}\Plus IP-Adapter~\cite{ipadapter}. }
\label{fig:fig_qual_1}
\end{figure*}

\begin{figure*}[!ht]
\centering
\includegraphics[width=1.0\linewidth]{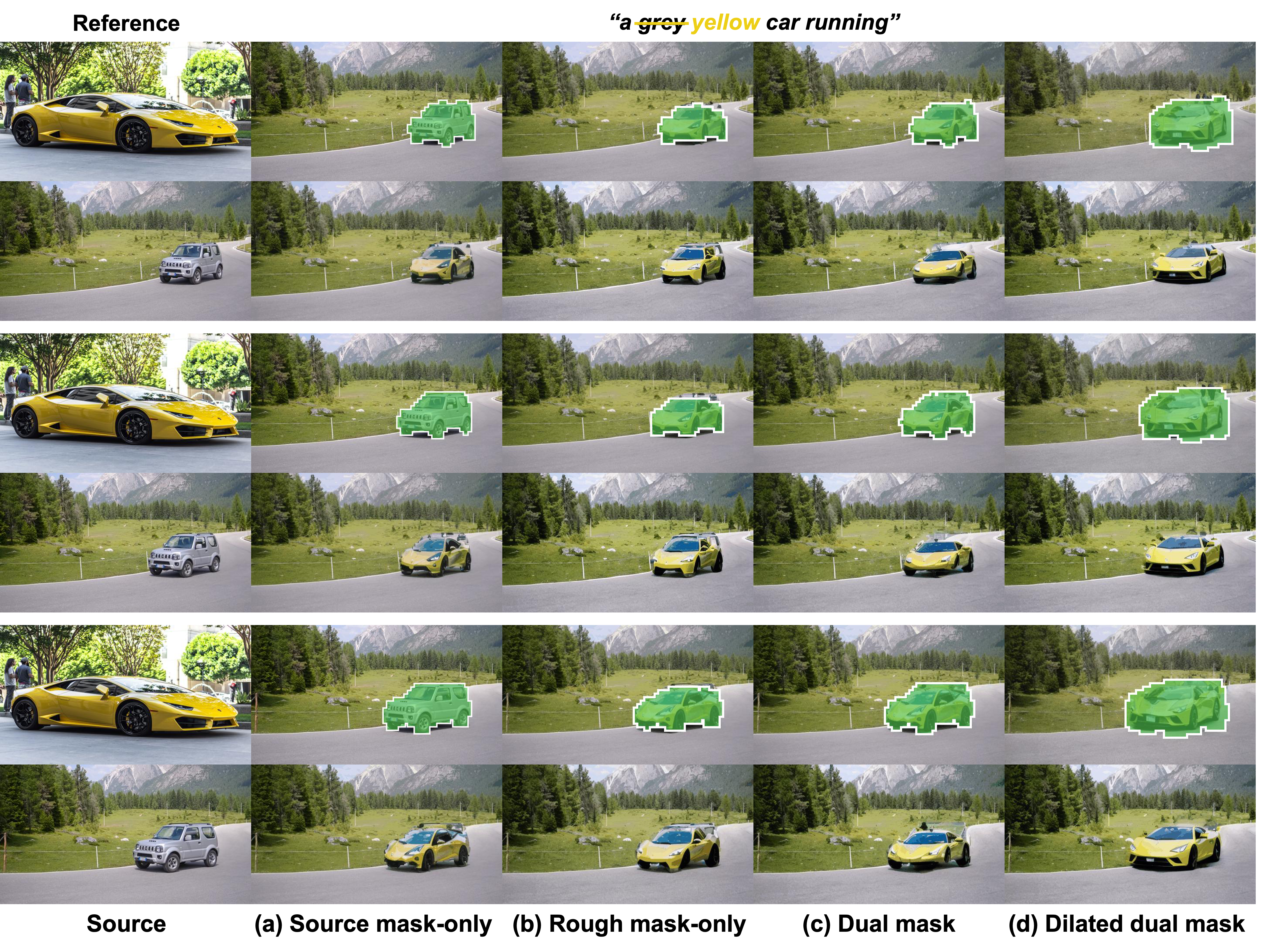}
\caption{\textbf{Full qualitative results of the ablation study on dilated dual masking.} Additional frames corresponding to Fig.~\ref{fig:ablation_mask} in the main manuscript are provided to further illustrate the consistent transformation of the object across time.}
\label{fig:add_frames}
\end{figure*}

\section{Full qualitative comparison and ablations}
\label{appen:frames}
This section presents the full qualitative comparisons with baseline methods and qualitative ablation results that complement those in the main paper. Appendix~\ref{appen:qual_comp} includes complete qualitative comparison against baseline approaches, Appendix~\ref{appen:abl} provides full results for ablation studies.

\begin{table}[!t]
\centering
\caption{\textbf{Quantitative evaluation of applicability to various T2V models.} We integrate our proposed components into AnimateDiff and TokenFlow. The integration improves reference alignment and image quality with minimal trade-offs across different architectures.} \label{table:supple_plug}
\resizebox{1.0\textwidth}{!}{%
\begin{tabular}{lcccc} 
\toprule
{\multirow{2.4}{*}{{\textbf{Method}}}} & \multicolumn{4}{c}{\textbf{Model scoring metrics}} \\
\cmidrule(lr){2-5}
& DINO~$\uparrow$ & CLIP-T~$\uparrow$ & Temp~$\uparrow$ & PSNR~$\uparrow$ \\
\midrule
AnimateDiff~\cite{animatediff} & 0.412 & 0.308 & \textbf{0.980} & 13.9 \\
  + \textbf{Ours} & \textbf{0.498}&\textbf{0.321}&0.972&\textbf{24.7}\\
\midrule
TokenFlow~\cite{tokenflow} &0.480&\textbf{0.325}&\textbf{0.975}& 19.6 \\
+ \textbf{Ours} & \textbf{0.523} & 0.320 & 0.974 & \textbf{23.9} \\
\bottomrule
\end{tabular}
}
\end{table}

\subsection{Qualitative comparison with baselines}\label{appen:qual_comp}
As shown in Fig.~\ref{fig:fig_qual_1}, since SDEdit-Video~\cite{sdedit} and AnimateDiff~\cite{animatediff}\Plus IP-Adapter~\cite{ipadapter} are adapted or composed from methods not originally designed for video editing, their qualitative results exhibit reduced visual fidelity compared to other baseline methods. Specifically, SDEdit-Video is extended from its original image-based formulation to the video domain, while AnimateDiff\Plus IP-Adapter combines AnimateDiff with IP-Adapter in a naive manner. As a result, SDEdit-Video preserves only the coarse spatial layout of the source video and often introduces excessive noise, leading to degraded visual quality. In the case of AnimateDiff\Plus IP-Adapter, reference images are directly pasted into the edited regions without sufficient consideration of semantic structure, resulting in visually unnatural artifacts.

\subsection{Qualitative ablation studies}\label{appen:abl}
\textbf{Dilated dual masking.}
Dilated dual masking enables the seamless replacement of the source object with the target object in the reference image across frames. Fig.~\ref{fig:add_frames} presents the complete results of Fig.~\ref{fig:ablation_mask}, confirming that the patterns observed remain consistent across different frames.

In Fig.~\ref{fig:add_frames}(a), when only the source mask is used, the target object is unnaturally constrained to the shape of the source mask, resulting in distortion. In Fig.~\ref{fig:add_frames}(b), using only the rough mask fails to fully cover the source object, leaving undesired remnants of it in the edited video. In Fig.~\ref{fig:add_frames}(c), employing a dual mask obtained by taking the union of the source and target masks, alleviates artifacts from the source object and shape mismatches of the target object, but the mask boundaries still lack smoothness. In contrast, our dilated dual mask in Fig.~\ref{fig:add_frames}(d) effectively mitigates these, enabling clean removal of the source object and coherent synthesis of the target object throughout the video. 

\begin{figure*}[!h]
\centering
\includegraphics[width=1.0\linewidth]{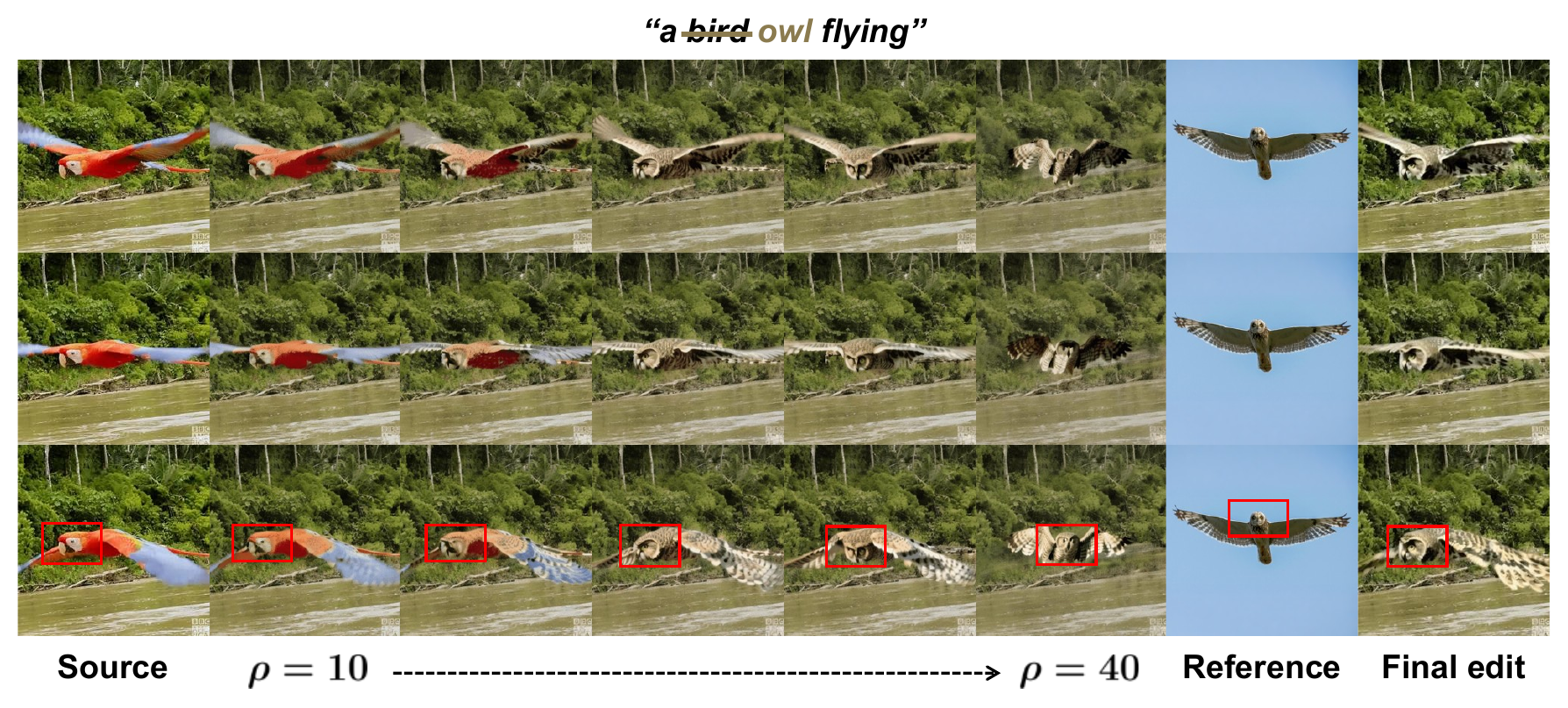}
\caption{\textbf{Full qualitative results of the ablation study on the $\rho$-start sampling.} We present the complete results corresponding to Fig.~\ref{fig:clean_latent} in the main manuscript by adding final edited results. As the value of \(\rho\)\ increases from 10 to 40 in Stage 1, the parrot in the source video gradually transforms to more closely resemble the owl in the reference image. After processing through Stage 2, the final edits further align with the reference appearance and exhibit improved visual quality.}
\label{supple_fig:rho}
\end{figure*}

\noindent
\textbf{$\rho$-start sampling.}
Our $\rho$-start sampling initiates DDIM sampling and inversion from intermediate time steps, thereby balancing the appearance of the reference image with the spatial layout of the source video. Fig.~\ref{supple_fig:rho} supplements Fig.~\ref{fig:clean_latent} by additionally presenting the final edited video results. Specifically, while Fig.~\ref{fig:clean_latent} only illustrates the first stage edits under varying values of $\rho$, the rightmost column in Fig.~\ref{supple_fig:rho} provides the corresponding final outputs after the second stage. The second stage further refines facial details, producing results that more closely resemble the reference image.

\begin{figure*}[!h]
\centering
\includegraphics[width=1.0\linewidth]{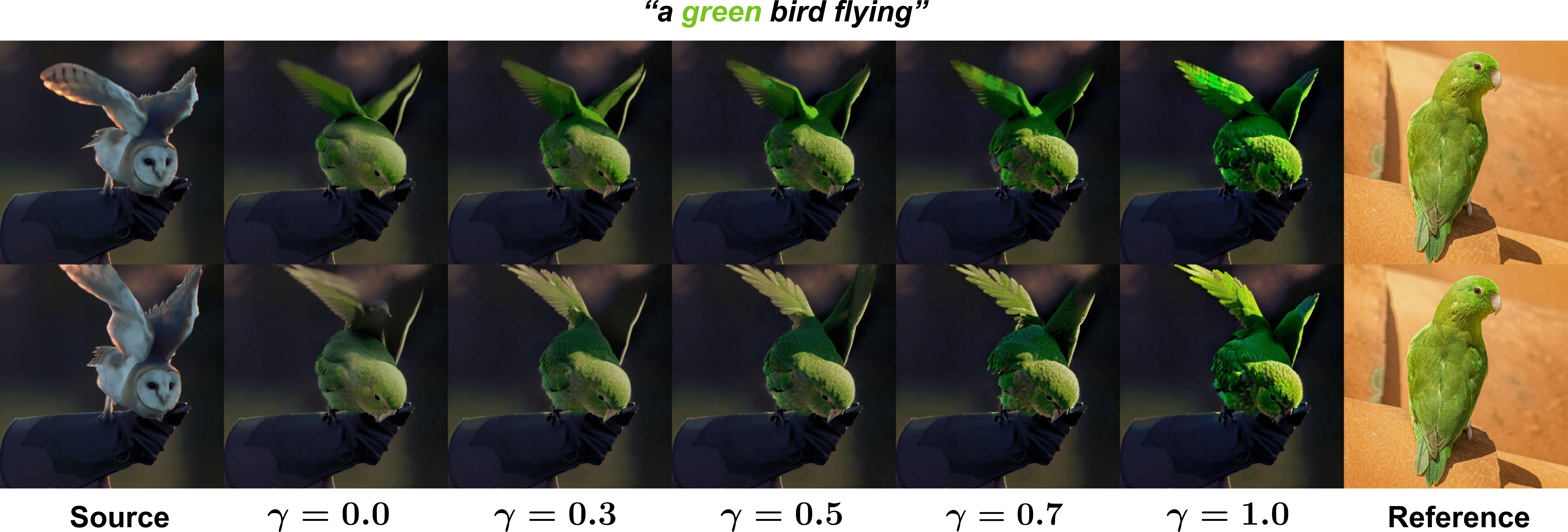}
\caption{\textbf{Full qualitative results of the ablation study on the zero image guidance scale \(\gamma\)\ .} We include the complete results corresponding to Fig.~\ref{fig:ablation_gamma} in the main paper. As the \(\gamma\)\ value is adjusted, the color fidelity of the edited video is calibrated to better match the reference image.}
\label{supple_fig:gamma}
\end{figure*}

\noindent
\textbf{Zero image guidance.}
Zero image guidance is introduced specifically to handle the occasional color desaturation observed in the edited videos. By adjusting the negative prompt to steer the generation away from the zero image direction, this mechanism effectively enhances color fidelity. Figure~\ref{supple_fig:gamma} provides an extension of the results in Figure~\ref{fig:ablation_gamma} by presenting edits across a wider range of $\gamma$ values. As $\gamma$ approaches 1, the contribution of zero image guidance within the negative image guidance increases, leading to noticeably improved color saturation in the edited videos.

\begin{figure*}[!t]
\centering  
\includegraphics[width=1.0\linewidth]{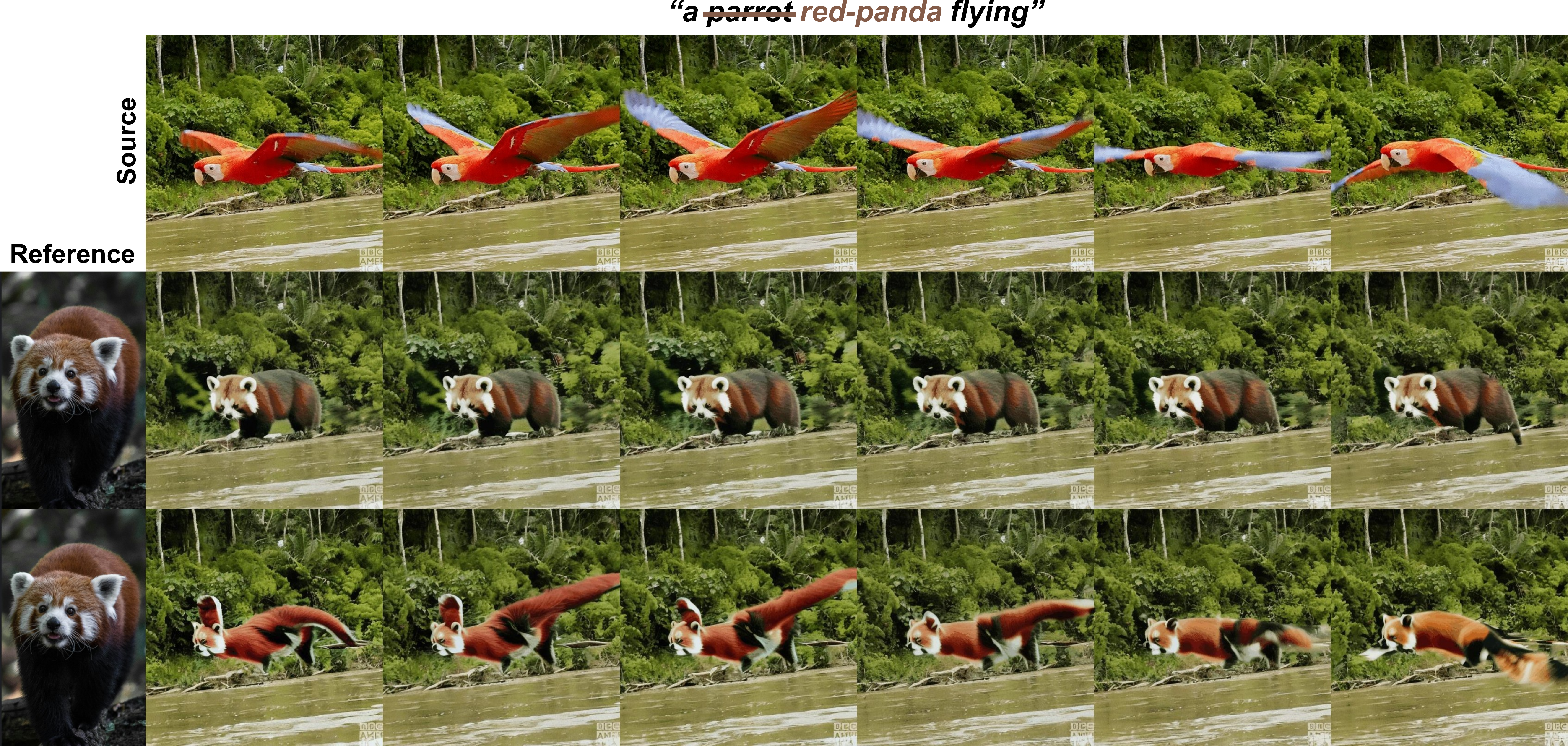}
\caption{\textbf{Example of failure case.} A large semantic gap between the object in the source video and the reference image poses challenges for our method. We show results from two versions of our method, obtained by varying the sampling start point $\rho$ and the dilation kernel size. In this case, editing a bird into a red panda results in missing or incorrect transfer of the wings and arms.}
\label{fig:limitation}
\end{figure*}

\section{Related work}
\label{appen:relatedwork}
\subsection{Image and video generation}
\label{appen:related_work_1}
Recent advancements in latent diffusion models (LDMs) have significantly improved the fidelity and controllability of image generation. Text-to-image (T2I) models such as Stable Diffusion~\cite{stablediffusion}, DALL-E 2~\cite{dalle2}, and Imagen~\cite{imagen} achieve high-resolution image synthesis aligned with complex text prompts. These models operate in the latent space for computational efficiency and support diverse conditioning mechanisms, including text, layout, and edge maps~\cite{ipadapter, controlnet}. Their flexibility has made LDMs foundational components for extending generative modeling into the video domain.

Motivated by T2I success, text-to-video (T2V) models have emerged to generate coherent video sequences conditioned on textual descriptions. CogVideo~\cite{cogvideo} adapts T2I backbones with temporal attention, enabling efficient video generation with minimal overhead. ZeroScope~\cite{zeroscope}, built on ModelScope~\cite{modelscope}, improves visual fidelity and consistency, producing higher-resolution outputs while mitigating common artifacts in earlier methods. AnimateDiff~\cite{animatediff} integrates temporal modules into Stable Diffusion~\cite{stablediffusion}, facilitating the T2V model with improved coherence between frames.

Alongside architectural advances, controllability and identity preservation have gained importance. Several methods leverage a reference frame along with a prompt to guide spatial appearance and layout across time~\cite{stiv, ti2v, controlvideo}. In addition, others introduce subject-driven personalization via instance finetuning~\cite{videoalchemist, dreamvideo, tune_a_video, customcrafter}, enabling video generation with identity consistency across diverse prompts. These trends illustrate a broader shift toward multimodal conditioning across text, image, and motion to allow more expressive and controllable video synthesis.

\subsection{Image and video editing}
\label{appen:related_work_2}
Following the advances in image and video generation models, diffusion models have also made significant progress in image and video editing, allowing precise content manipulation. In the image domain, Prompt-to-Prompt~\cite{prompt-to-prompt} introduces cross-attention manipulation within pretrained diffusion models, enabling semantic edits without finetuning or masks. InstructPix2Pix~\cite{instructpix2pix} leverages pairs of instructions and images to support editing guided by natural language. EditAnything~\cite{editanything} generalizes this paradigm by integrating multi-modal inputs such as text, masks, and user clicks.

Video editing extends these capabilities to the temporal dimension. Tune-A-Video~\cite{tune_a_video} employs one-shot tuning of pretrained image diffusion models for video editing tasks, allowing users to modify video content and style with minimal data and strong temporal consistency. Video-P2P~\cite{videop2p} extends cross-attention control techniques from image to video, maintaining both spatial and temporal coherence. Text2Video-Zero~\cite{text2video} removes the need for paired text-video data, instead leveraging pretrained text-image models for flexible and data-efficient video editing.

To overcome the limitations of editing based solely on text, recent studies propose image-conditioned video editing methods that use a reference image to guide appearance and identity~\cite{genvideo, make_a_pro, cutandpaste}. Make-A-Protagonist~\cite{make_a_pro} introduces a unified framework for protagonist replacement, background editing, and style transfer using mask-guided denoising. GenVideo~\cite{genvideo} proposes shape-aware masking and latent noise correction to ensure temporal consistency and subject fidelity. VideoSwap~\cite{videoswap} adds semantic point correspondences to provide interactive and precise control over the subject’s motion and appearance. Despite these advancements, existing methods still require finetuning on each source video and rely on user-defined semantic keypoints, limiting practicality and scalability. In contrast, our method directly utilizes the image prompt without any finetuning, while effectively preserving temporal consistency and semantic accuracy.

\section{Limitation}\label{appen:limit}
Although our method exhibits notable improvements over existing baselines, as evidenced by extensive experiments, it still faces certain limitations. When there is a substantial disparity in the innate physical characteristics between the source and target objects, our method struggles to perform object replacement. Figure~\ref{fig:limitation} shows that our method generates wings for the red panda which is an anatomically implausible feature to transfer the parrot’s flying motion. Conversely, in attempting to preserve the red panda’s natural characteristics, it fails to convey the intended motion.

\section{Additional experimental settings}\label{appen:exp_settings}

Appendix~\ref{appen:implement} provides further implementation details, including hyperparameter settings. Appendix~\ref{appendix:reproduce} outlines the reproducibility settings for all baseline models used in our experiments.
Appendix~\ref{appen:userstudy} presents the evaluation questions used in our user study.

\subsection{Additional implementation details}\label{appen:implement}
Our method introduces four key hyperparameters: $\rho$, $\gamma$, dilation kernel size, and guidance scale. Each parameter has a well-defined and narrow range that yields consistent performance across diverse editing tasks. Consequently, hyperparameter tuning in the proposed method requires minimal manual adjustment and can be efficiently performed without extensive search or optimization.

\begin{itemize}
    \item $\rho \in [10, 30]$: Controls the starting timestep for the denoising trajectory. In practice, one of the discrete values (10, 20, or 30) generally performs well. Users may optionally refine the result by exploring intermediate values.
    
    \item $\gamma \in [0, 1]$: Controls the weighted combination of standard zero-vector and zero-image embeddings in zero image guidance. The default value of 0.5 has been empirically found to perform robustly across a wide variety of scenes, object types, and motion dynamics.
    
    \item dilation kernel size \(k\) $\in [0, 10]$: Defines the kernel size used in the dilated dual masking. It affects the spatial smoothness and coverage of the edited region. The default value of 3 works well in most cases, while slightly larger values (e.g., 5 or 7) can help when the source and target objects differ significantly in shape or size.
    
    \item guidance scale $\in [6, 10]$: Determines the strength of classifier-free guidance during denoising. The default setting of 6 typically yields semantically faithful results. 
\end{itemize}
Depending on the resolution of the source video, we use one of the following sizes: 320×576, 432×768, 480×480, 515×512, 600×600, 640×640, or 700×700. As an exception, the results in Fig.~\ref{fig:teaser} are rendered at 1024×576 to better showcase the our method's performance. The number of frames is set to 8 or 16, selected according to the source video length.

\subsection{Reproducibility of baseline models}~\label{appendix:reproduce}
All baseline models are run using the official implementations released in their GitHub repositories. For a fair comparison, we evaluate all baselines under the same experimental settings as our method unless specified otherwise (refer to Appendix~\ref{appen:implement}). The same text prompts are applied across all methods, except for the method that follows its own text prompting strategy. We also use the same spatial resolution for \ours and all baseline models, except for methods that inherently require square inputs.  

{\setcounter{footnote}{0}
\renewcommand{\thefootnote}{\arabic{footnote}}%
\begin{itemize}
    \item \textbf{SDEdit-Video}~\cite{sdedit}\footnote{\url{https://github.com/huggingface/diffusers/blob/main/src/diffusers/pipelines/animatediff/pipeline_animatediff_video2video.py}}
        We adopt the video-extended implementation provided by HuggingFace, as the original SDEdit is originally image-based. 
            \item \textbf{FateZero}~\cite{fatezero}\footnote{\url{https://github.com/ChenyangQiQi/FateZero}}
    As FateZero~\cite{fatezero} supports only square outputs, we paste its generated video into the corresponding region of the source video for comparison. Both FateZero~\cite{fatezero} and Make-A-Pro~\cite{make_a_pro} employ model-based prompting procedures to automatically generate text prompts. While FateZero originally uses an earlier BLIP variant for this step, the overall prompting procedure is fundamentally the same as that of Make-A-Protagonist, which uses BLIP-2. To maintain consistency, we adopt the more recent BLIP-2–based prompting strategy for FateZero as well and use identical text prompts for both methods.
            \item \textbf{Video-P2P}~\cite{videop2p}\footnote{\url{https://github.com/dvlab-research/Video-P2P}}
        Similar to FateZero, since Video-P2P~\cite{videop2p} produces square outputs, we paste its generated video into the corresponding region of the source video.
            \item \textbf{TokenFlow}~\cite{tokenflow}\footnote{\url{https://github.com/omerbt/TokenFlow}}
        TokenFlow is evaluated under the same experimental settings as our method without any additional modifications. To implement \textbf{TokenFlow+Ours}, we extend TokenFlow by adding IP-Adapter~\cite{ipadapter} to support image conditioning, and incorporate our proposed approaches, including zero-image guidance, $\rho$-start sampling, and dilated dual masking. 
            \item \textbf{AniamteDiff}~\cite{animatediff}\footnote{\url{https://github.com/guoyww/AnimateDiff}}\textbf{\Plus IP-Adapter}~\cite{ipadapter}\footnote{\url{https://github.com/tencent-ailab/IP-Adapter}}
     For both AnimateDiff and IP-Adapter, we use their official GitHub implementations, and fuse the two following the integration scheme proposed in the IP-Adapter paper.
    \item \textbf{Make-A-Protagonist}~\cite{make_a_pro}\footnote{\url{https://github.com/HeliosZhao/Make-A-Protagonist}}
        Following the original prompt generation strategy, we extract protagonist cues using BLIP-2~\cite{blip2} and generate prompts via visual question answering as provided in its official GitHub repository.
\end{itemize}
}
\subsection{User study details}
\label{appen:userstudy}
In the user study described in Section~\ref{sec:experiment}, participants were asked to rank the seven generated videos from 1st (best) to 7th (worst) for each of the following five evaluation criteria:
\begin{itemize}
    \item \textbf{Image Alignment}: How well the appearance of the "object in the generated video" matches the "object in the reference image (excluding the background)".

    \item \textbf{Text Alignment}: How well "the content of the generated video" aligns with the "provided text prompt".  
    
    \textit{Text prompt:} ``a polar bear skiing''
    
    \item \textbf{Motion Alignment}: How closely the "motion of the object in the generated video" matches the "motion of the object in the source video".

    \item \textbf{Temporal Consistency}: The degree to which the object and background remain consistent  across time.

    \item \textbf{Overall Quality}: The overall realism and naturalness of the edited video, considering all aspects above.
\end{itemize}

\section{Algorithm}\label{appen:alg}
The overall video editing pipeline of \ours is outlined in Algorithm~\ref{algorithm}. Our training-free method consists of two stages: a rough editing stage that generates an initial object transformation guided by the source mask, and a refinement stage that further improves visual fidelity and temporal coherence using an updated mask derived from the intermediate result.

\begin{algorithm*}
\caption{Training-free video editing procedure of \textbf{\ours}}\label{algorithm}
\begin{algorithmic}[1]

\Statex  \textbf{Input:} Source video $\mathcal{V}_s$, image prompt $\mathcal{R}$, text prompt $\mathcal{T}$, negative text prompt $\bar{\mathcal{T}}$
\Statex  \textbf{Output:} Edited video $\mathcal{V}_e$

\Statex \gbbox[0.985]{\textit{Prompt feature extraction}}
\State $\vc_\text{img} \gets \text{CLIP}_\text{img}(\mathcal{R}), \vc_\text{txt} \gets \text{CLIP}_\text{txt}(\mathcal{T})$ 

\State $\bar{\vc}_\text{img} = \gamma \cdot \text{CLIP}_\text{img}(\rmI_0) + (1-\gamma) \cdot \mathbf{0}$, $\bar{\vc}_\text{txt} = \gamma \cdot \text{CLIP}_\text{txt}({\mathcal{\bar{T}}}), \text{ where } \gamma \in [0,1]$

\Statex  \lbbox[0.985]{\textit{\textbf{Stage 1: Rough editing}}}

\Statex   \gbbox[0.985] {\textit{Dilated mask generation}}
\State  $\mM_{s} \gets \text{Segment}(\mathcal{V}_s)$ 
\State  $\mM'_{s} \gets \text{Dilate}(\mM_{s})$ 

\Statex   \gbbox[0.985]{\textit{Latent encoding}}
\State  $\vz_0 \gets \mathcal{E}(\mathcal{V}_s)$ 
\State  $\{ \vz_t \}_{t = \tau_1}^{\tau_\nu} \gets \text{DDIM-inv}(\vz_0)$ 

\Statex   \gbbox[0.985]{\textit{Denoising}}

\State  Select $\rho \in [\nu]$ 
\State  $\vz'_{\tau_\rho} \gets \sqrt{\alpha_{\tau_\rho}} \cdot \vz_0 + \sqrt{1 - \alpha_{\tau_\rho}} \cdot \eta$, where $\eta \sim \mathcal{N}(0, \mI)$ 
\For{$t = \rho$ to $1$}
    \State  $\fgt_{\tau_{t\Minus1}} \gets \text{DDIM-samp}(\vz'_{\tau_t}, \vc_\text{img}, \vc_\text{txt}, \bar{\vc}_\text{img}, \bar{\vc}_\text{txt}, \tau_t)$ 
    \State  $\bgt_{\tau_{t\Minus1}} \gets \vz_{\tau_{t\Minus1}}$ 
    \State  $\vz'_{\tau_{t\Minus1}} \gets \fgt_{\tau_{t\Minus1}} \odot \mM'_{s} + \bgt_{\tau_{t\Minus1}} \odot (1 - \mM'_{s})$ 
 \EndFor

\Statex \gbbox[0.985]{\textit{Latent decoding}}
\State  $\mathcal{V}_r \gets \mathcal{D}(\vz'_{\tau_0})$ 
\State  $\mathcal{V}_r \gets \mathcal{V}_r \odot \mM'_s +  \mathcal{V}_s \odot (1-\mM'_s)$

\Statex \lbbox[0.985]{\textit{\textbf{Stage 2: Refined editing}}}

\Statex \gbbox[0.985]{\textit{Dilated dual mask generation}}
\State  $\mM_{r} \gets \text{Segment}(\mathcal{V}_r)$ 
\State  $\mM'_{r} \gets \text{Dilate}(\mM_{r})$ 
\State  $\mM_{e} \gets \mM'_s \cup \mM'_r $ 

\Statex \gbbox[0.985]{\textit{Latent encoding}}
\State  $\vz_0 \gets \mathcal{E}(\mathcal{V}_r)$ 
\State  $\{ \vz_t \}_{t = \tau_1}^{\tau_\nu} \gets \text{DDIM-inv}(\mathcal{E}(\mathcal{V}_s))$ 

\Statex \gbbox[0.985]{\textit{Denoising}}

\State  Select $\rho \in [\nu]$ 
\State  $\vz'_{\tau_\rho} \gets \sqrt{\alpha_{\tau_\rho}} \cdot \vz_0 + \sqrt{1 - \alpha_{\tau_\rho}} \cdot \eta$, where $\eta \sim \mathcal{N}(0, \mI)$ 
 \For{$t = \rho$ to $1$}
     \State  $\fgt_{\tau_{t\Minus1}} \gets \text{DDIM-samp}(\vz'_{\tau_t}, \vc_\text{img}, \vc_\text{txt}, \bar{\vc}_\text{img}, \bar{\vc}_\text{txt}, \tau_t)$ 
    \State  $\bgt_{\tau_{t\Minus1}} \gets \vz_{\tau_{t\Minus1}}$ 
    \State  $\vz'_{\tau_{t\Minus1}} \gets \fgt_{\tau_{t\Minus1}} \odot \mM_{e} + \bgt_{\tau_{t\Minus1}} \odot (1 - \mM_{e})$ 
 \EndFor

\Statex \gbbox[0.985]{\textit{Latent decoding}}
\State  $\mathcal{V}_e \gets \mathcal{D}(\vz'_{\tau_0})$  

\end{algorithmic}
\end{algorithm*}

\section{Additional qualitative results}\label{appen:qual}

\begin{figure*}[!t]
\centering  
\includegraphics[width=1.0\linewidth]{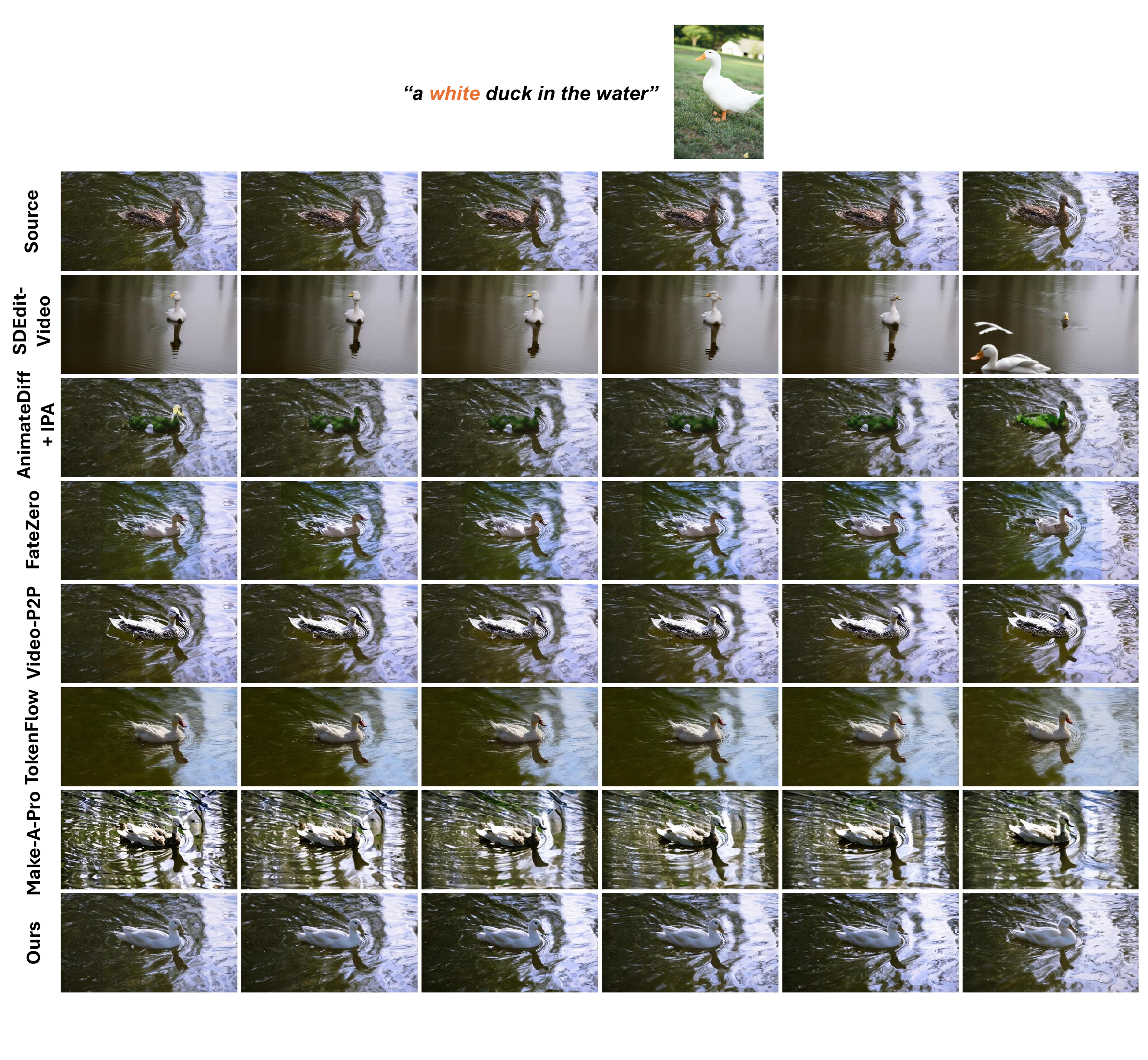}
\caption{\textbf{Complete video editing comparisons with baseline methods.} FateZero~\cite{fatezero} and Make-a-Pro~\cite{make_a_pro} use text prompts generated by BLIP-2~\cite{blip2}, following their respective proposed strategies. TokenFlow~\cite{tokenflow}, Video-P2P~\cite{videop2p}, FateZero~\cite{fatezero}, and SDEdit-Video~\cite{sdedit} use only text prompts, while Make-a-Pro~\cite{make_a_pro}, AnimateDiff~\cite{animatediff}+IP-Adapter~\cite{ipadapter}, and our proposed method use both image and text prompts.}
\label{fig:fig_qual_2}
\end{figure*}

\begin{figure*}[!t]
\centering  
\includegraphics[width=1.0\linewidth]{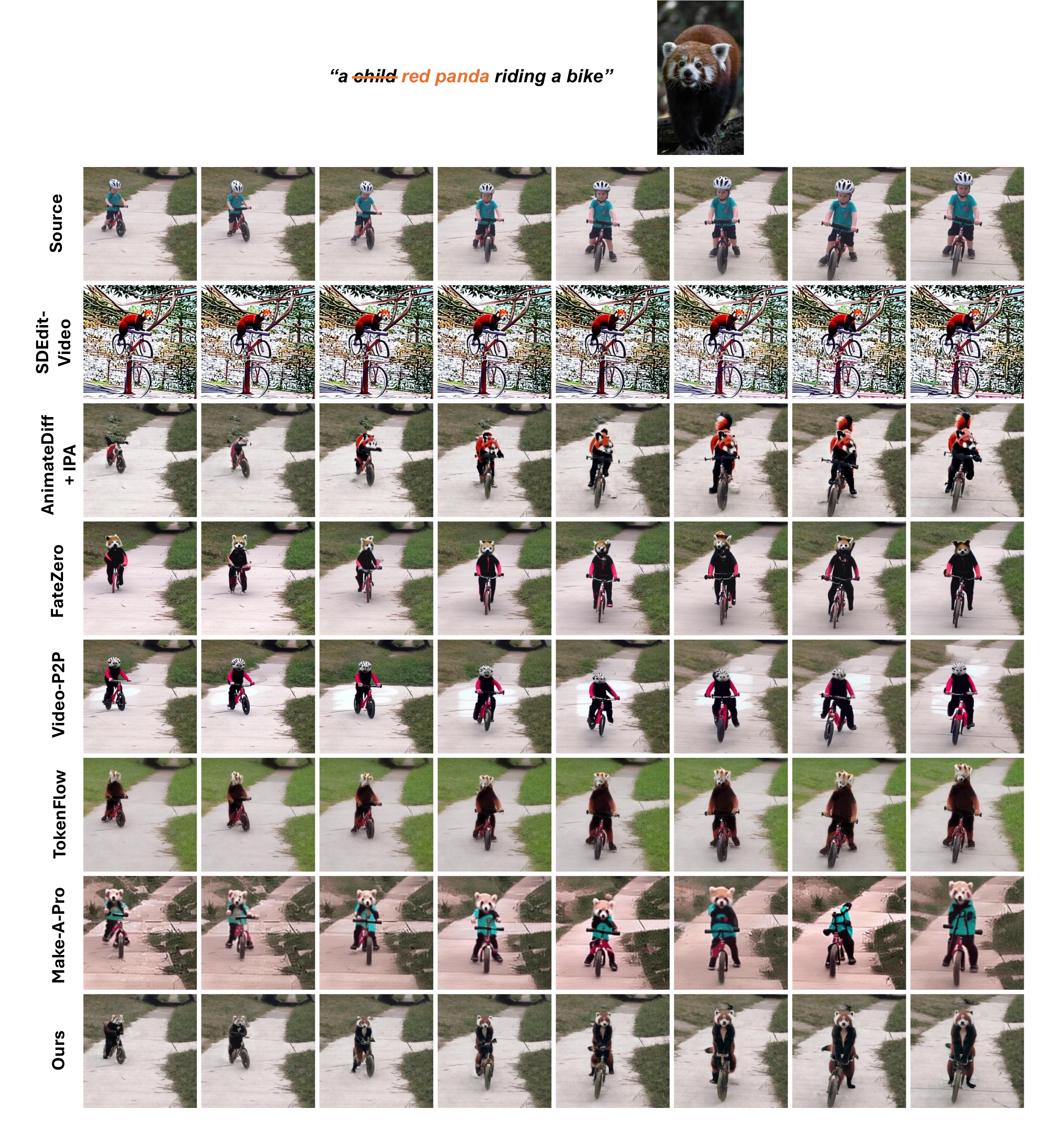} 
\caption{\textbf{Complete video editing comparisons with baseline methods.} FateZero~\cite{fatezero} and Make-a-Pro~\cite{make_a_pro} use text prompts generated by BLIP-2~\cite{blip2} following their respective proposed strategies. TokenFlow~\cite{tokenflow}, Video-P2P~\cite{videop2p}, FateZero~\cite{fatezero}, and SDEdit-Video~\cite{sdedit} use only text prompts, while Make-a-Pro~\cite{make_a_pro}, AnimateDiff~\cite{animatediff}+IP-Adapter~\cite{ipadapter}, and our proposed method use both image and text prompts.}
\label{fig:fig_qual_3}
\end{figure*}

\begin{figure*}[!t]
\centering  
\includegraphics[width=1.0\linewidth]{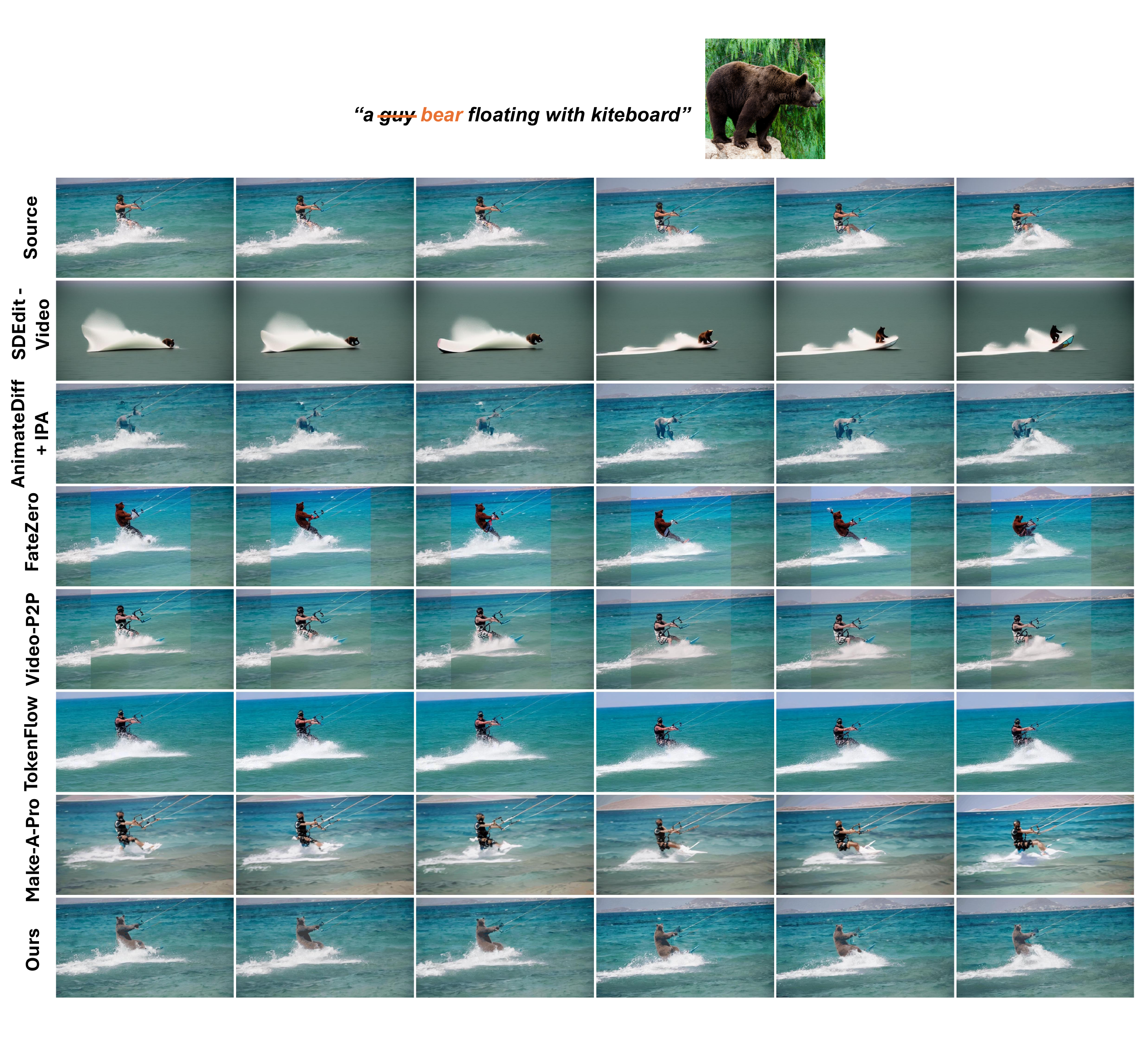} 
\caption{\textbf{Complete video editing comparisons with baseline methods.} FateZero~\cite{fatezero} and Make-a-Pro~\cite{make_a_pro} use text prompts generated by BLIP-2~\cite{blip2} following their respective proposed strategies. TokenFlow~\cite{tokenflow}, Video-P2P~\cite{videop2p}, FateZero~\cite{fatezero}, and SDEdit-Video~\cite{sdedit} use only text prompts, while Make-a-Pro~\cite{make_a_pro}, AnimateDiff~\cite{animatediff}+IP-Adapter~\cite{ipadapter}, and our proposed method use both image and text prompts.}
\label{fig:fig_qual_4}
\end{figure*}

\begin{figure*}[!t]
\centering  
\includegraphics[width=1.0\linewidth]{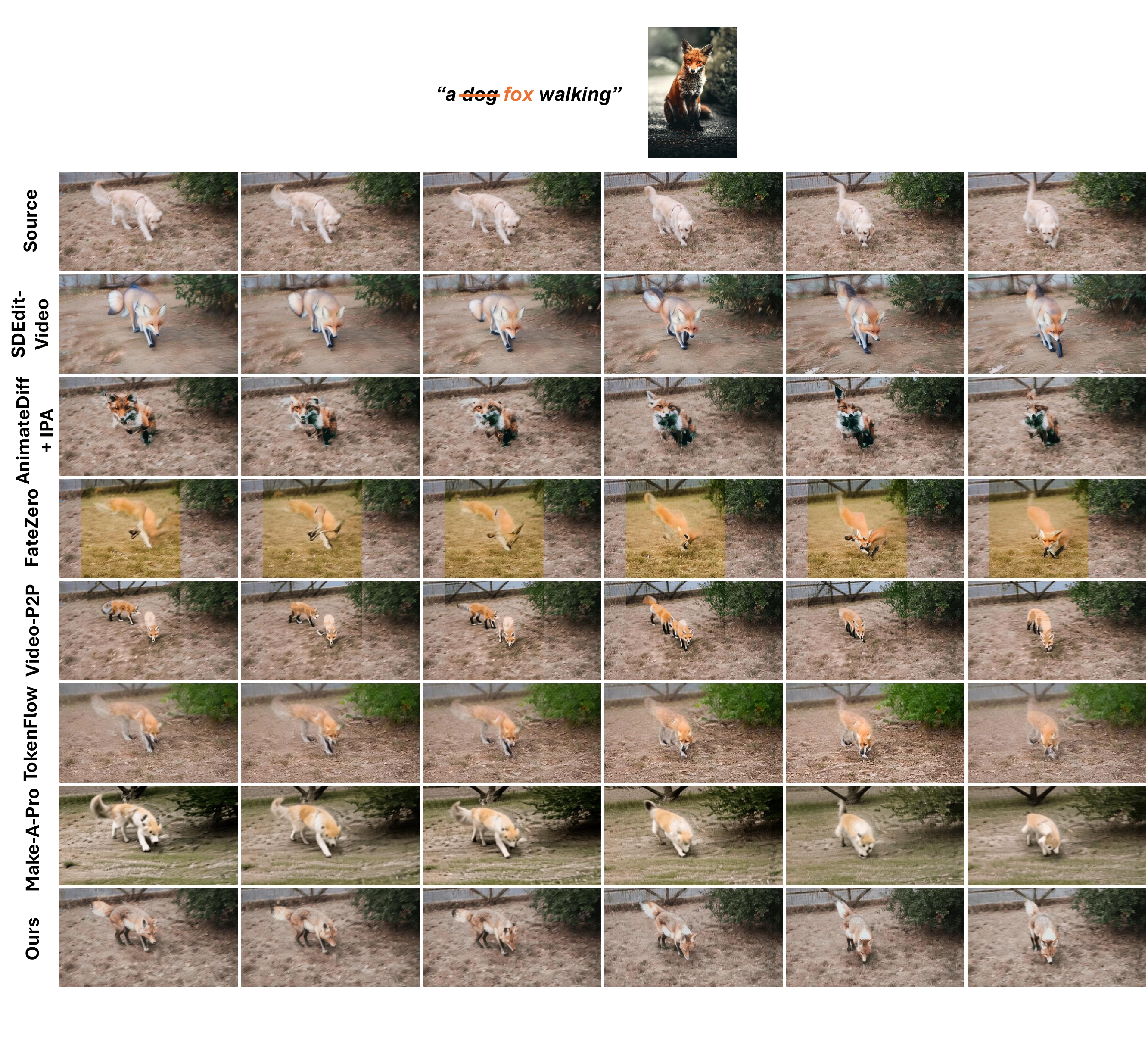} 
\caption{\textbf{Complete video editing comparisons with baseline methods.} FateZero~\cite{fatezero} and Make-a-Pro~\cite{make_a_pro} use text prompts generated by BLIP-2~\cite{blip2} following their respective proposed strategies. TokenFlow~\cite{tokenflow}, Video-P2P~\cite{videop2p}, FateZero~\cite{fatezero}, and SDEdit-Video~\cite{sdedit} use only text prompts, while Make-a-Pro~\cite{make_a_pro}, AnimateDiff~\cite{animatediff}+IP-Adapter~\cite{ipadapter}, and our proposed method use both image and text prompts.}
\label{fig:fig_qual_5}
\end{figure*}

\begin{figure*}[!t]
\centering  
\includegraphics[width=1.0\linewidth]{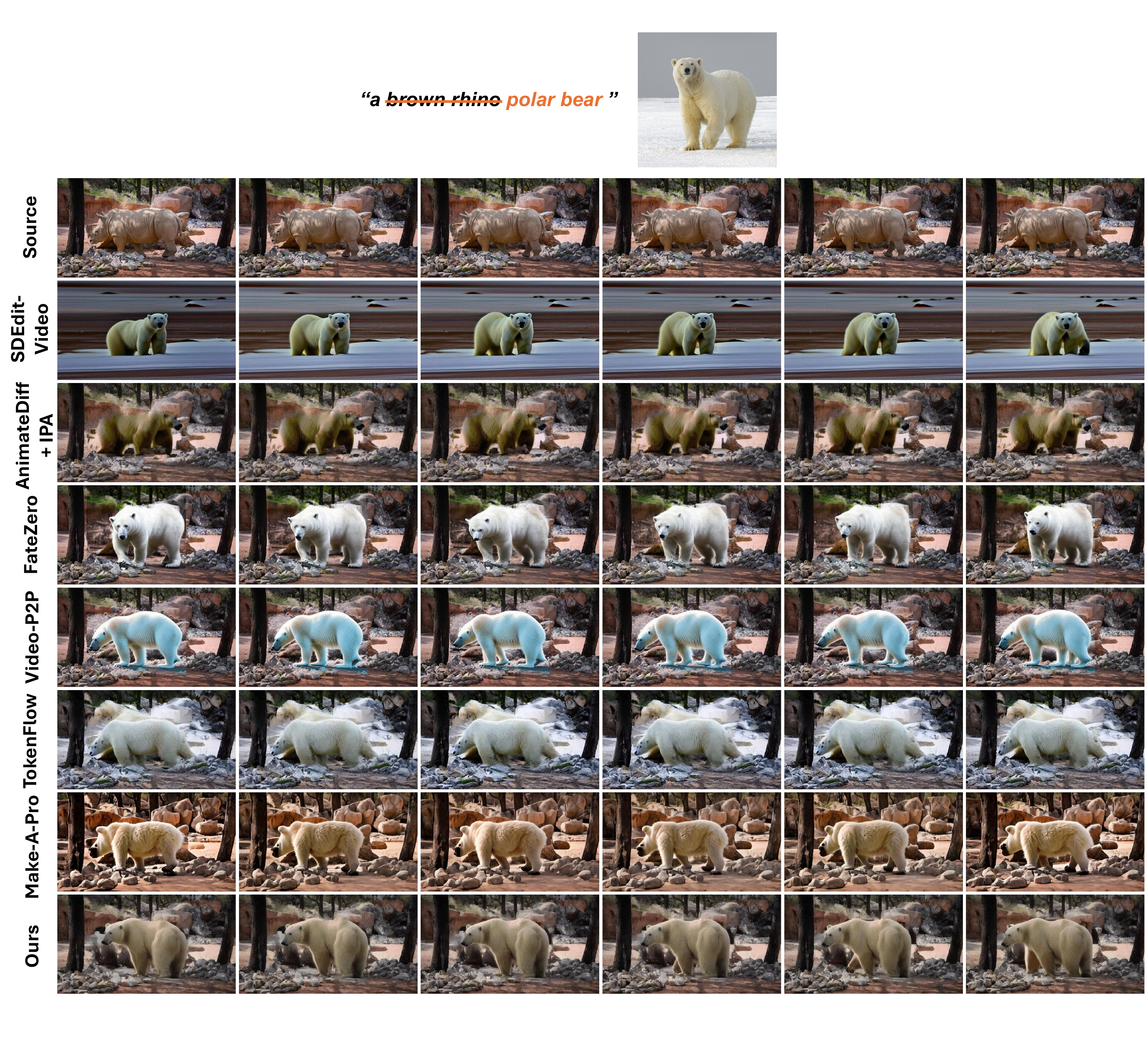} 
\caption{\textbf{Complete video editing comparisons with baseline methods.} FateZero~\cite{fatezero} and Make-a-Pro~\cite{make_a_pro} use text prompts generated by BLIP-2~\cite{blip2} following their respective proposed strategies. TokenFlow~\cite{tokenflow}, Video-P2P~\cite{videop2p}, FateZero~\cite{fatezero}, and SDEdit-Video~\cite{sdedit} use only text prompts, while Make-a-Pro~\cite{make_a_pro}, AnimateDiff~\cite{animatediff}+IP-Adapter~\cite{ipadapter}, and our proposed method use both image and text prompts.}
\label{fig:fig_qual_7}
\end{figure*}

Additional qualitative comparisons with baseline methods are provided in Fig.~\ref{fig:fig_qual_2}, Fig.~\ref{fig:fig_qual_3}, Fig.~\ref{fig:fig_qual_4}, Fig.~\ref{fig:fig_qual_5}, and Fig.~\ref{fig:fig_qual_7}, which further demonstrate the superior visual fidelity and temporal consistency achieved by our approach. 
Notably, AnimateDiff+IP-Adapter frequently exhibits noticeable distortions in both shape and color, leading to unnatural visual artifacts. To address this, our method constructs a well-formed noise maps by selecting a less-degraded latent via \(\rho\)-start sampling and combining foreground and background latents through dilated dual masking. This structured noise formulation enables accurate object replacement while preserving the motion dynamics of the source video.

\end{document}